\def\year{2022}\relax
\documentclass[letterpaper]{article} 
\usepackage{aaai22}  
\usepackage{times}  
\usepackage{helvet}  
\usepackage{courier}  
\usepackage[hyphens]{url}  
\usepackage{graphicx} 
\def\UrlFont{\rm}  
\usepackage{natbib}  
\usepackage{caption} 
\DeclareCaptionStyle{ruled}{labelfont=normalfont,labelsep=colon,strut=off} 
\usepackage{algorithm}
\usepackage{algorithmic}

\usepackage{newfloat}
\usepackage{listings}
\floatstyle{ruled}
\newfloat{listing}{tb}{lst}{}
\floatname{listing}{Listing}
\usepackage{url}            
\usepackage{booktabs}       
\usepackage{amsfonts}       
\usepackage{amssymb}
\usepackage{amsmath}
\usepackage{nicefrac}       
\usepackage{microtype}      
\usepackage{xcolor}         
\usepackage{bm}
\usepackage{graphicx}
\usepackage{caption}
\usepackage{subcaption}
\usepackage{xspace}
\usepackage{cleveref}

\newcommand{\vect}[1]{\boldsymbol{#1}}

\newcommand{\psgi}{PSGI\xspace}

\newcommand{\FG}{\mathcal{G}}
\newcommand{\FGpcond}{\mathcal{G}_\text{prec}}
\newcommand{\FGeffect}{\mathcal{G}_\text{eff}}
\newcommand{\FGr}{\mathcal{G}_r}
\newcommand{\hatFGpcond}{\widehat{\mathcal{G}}_\text{prec}}
\newcommand{\hatFGeffect}{\widehat{\mathcal{G}}_\text{eff}}
\newcommand{\hatFGr}{\widehat{\mathcal{G}}_r}

\newcommand{\entityspace}{\mathcal{E}}
\newcommand{\entityembed}{f_\text{entityembed}}

\newcommand{\subtaskspace}{\Phi}
\newcommand{\subtask}{\phi}
\newcommand{\optionspace}{\mathbb{O}}
\newcommand{\option}{\mathcal{O}}

\def\mine{\textbf{Mining}\xspace}
\def\cook{\textbf{Cooking}\xspace}
\def\thor{\textbf{AI2Thor}\xspace}

\newcommand{\fsgibf}{\textbf{PSGI}\xspace}
\def\msgiorg{\textbf{MSGI}\xspace}
\def\msgi{\textbf{MSGI$^{+}$}\xspace}
\def\random{\textbf{Random}\xspace}

\def\hrl{\textbf{HRL}\xspace}

\input{math_commands}

\newcommand{\todo}[1]{}
\newcommand{\warning}[1]{}
\renewcommand\warning[1]{\textcolor{purple}{#1}}

\newcommand{\cutabstractup}{}
\newcommand{\cutabstractdown}{}

\newcommand{\cutsectionup}{}
\newcommand{\cutsectiondown}{}

\newcommand{\cutsubsectionup}{}
\newcommand{\cutsubsectiondown}{}

\newcommand{\cutparagraphup}{}

\title{Learning Parameterized Task Structure \\for Generalization to Unseen Entities}

\author {
    Anthony Liu\equalcontrib \textsuperscript{\rm 1},
    Sungryull Sohn\equalcontrib \textsuperscript{\rm 2},
    Mahdi Qazwini \textsuperscript{\rm 1 },
    Honglak Lee \textsuperscript{\rm 1 2}
}
\affiliations {
    \textsuperscript{\rm 1}
    University of Michigan\\
    \textsuperscript{\rm 2}
    LG AI Research\\
    anthliu@umich.edu, srsohn@lgresearch.ai, mqazwini@umich.edu, honglak@eecs.umich.edu
}

\begin{document}

\maketitle

\cutabstractup
\begin{abstract}
\cutabstractdown
Real world tasks are hierarchical and compositional. Tasks can be composed of multiple subtasks (or sub-goals) that are dependent on each other. These subtasks are defined in terms of entities (e.g., \texttt{apple, pear}) that can be recombined to form new subtasks (e.g., \texttt{pickup apple}, and \texttt{pickup pear}).
To solve these tasks efficiently, an agent must infer subtask dependencies (e.g. an agent must execute \texttt{pickup apple} before \texttt{place apple in pot}), and generalize the inferred dependencies to new subtasks (e.g. \texttt{place apple in pot} is similar to \texttt{place apple in pan}).
Moreover, an agent may also need to solve unseen tasks, which can involve unseen entities.
To this end, we formulate parameterized subtask graph inference (PSGI), a method for modeling subtask dependencies using first-order logic with subtask entities.
To facilitate this, we learn entity attributes in a zero-shot manner, which are used as quantifiers (e.g. \texttt{is\_pickable(X)}) for the parameterized subtask graph. 
We show this approach accurately learns the latent structure on hierarchical and compositional tasks more efficiently than prior work, and show PSGI can generalize by modelling structure on subtasks unseen during adaptation\footnote{Code is available at \url{https://github.com/anthliu/PSGI}}.

\end{abstract}
\cutsectionup
\section{Introduction}
\begin{figure*}
  \centering
  \includegraphics[width=0.85\textwidth]{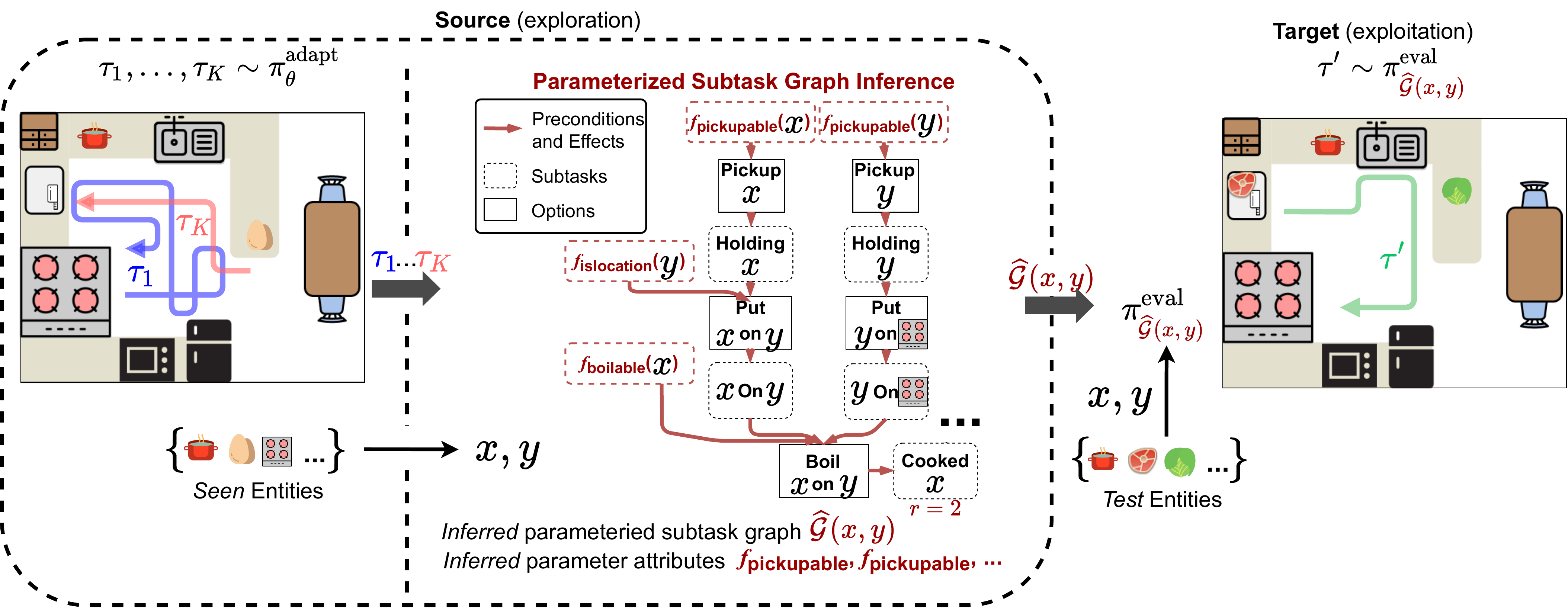}
  \vspace{-5pt}
  \caption{
  We present an overview of \textit{parameterized subtask graph inference} (\psgi)
  in a toy cooking environment. In various tasks, the agent must cook various foods to receive reward.
  \textbf{Left:} The adaptation policy \smash{$\pi_\theta^\text{adapt}$} initially explores
  the cooking source task (training), generating a trajectories \smash{$\tau_1, \dots, \tau_K$}.
  \textbf{Middle:} Using \smash{$\tau_1,\dots, \tau_K$}, the agent infers a \textit{parameterized subtask graph} $\widehat{\FG}$ of the environment, which describes the preconditions
  and effects between \textit{parameterized} options and subtasks using ($x$ and $y$) over
  \textit{entities} (objects in the environment).
  The agent learns a set of parameter attributes \smash{($\widehat{A}_\text{att} = f_\text{pickupable}\dots$)}
  in a \textit{zero-shot} manner and uses these attributes to construct $\widehat{\FG}$.
  \textbf{Right:} The agent initializes a separate test policy $\pi_{\widehat{\FG}}^\text{test}$ that maximizes reward by following the inferred
  parameterized subtask graph \smash{$\widehat{\FG}$}. In this target environment (test) there exist
  \textit{unseen parameters} (cabbage and meat). Preconditions and effects for these
  parameters are accurately inferred by substituting for entities ($x$ and $y$).
  }
  \vspace{-10pt}
  \label{fig:psgi-overview}
\end{figure*}

\cutsectiondown
Real world tasks are \textit{hierarchical}.
Hierarchical tasks are composed of multiple sub-goals that must be completed
in certain order. For example, the cooking task shown in Figure~\ref{fig:psgi-overview}
requires an agent to boil some food object (e.g.\ \texttt{Cooked egg}).
An agent must place the food object $x$ in a cookware object $y$, place the cookware object
on the stove, before boiling this food object $x$. Parts of this task
can be decomposed into sub-goals, or \textit{subtasks} (e.g.\ \texttt{Pickup egg}, \texttt{Put egg on pot}).
Solving these tasks requires long horizon planning and reasoning ability~\cite{erol1996hierarchical,xu2018neural,ghazanfari2017autonomous,sohn2018hierarchical}.
This problem is made more difficult of rewards are \textit{sparse}, if only
few of the subtasks in the environment provide reward to the agent.

Real world tasks are also \textit{compositional}~\cite{carvalho2020reinforcement,loula2018rearranging,andreas2017modular,oh2017zero}.
Compositional tasks are often made of different ``components" that can recombined
to form new tasks. These components can be numerous, leading to a \textit{combinatorial}
number of subtasks. For example, the cooking task shown in Figure~\ref{fig:psgi-overview}
contains subtasks that follow a \text{verb-objects} structure. The verb \texttt{Pickup}
admits many subtasks, where any object $x$ composes into a new subtask (e.g.\ \texttt{Pickup egg}, \texttt{Pickup pot}).
Solving compositional tasks also requires reasoning~\cite{andreas2017modular,oh2017zero}.
Without reasoning on the relations between components between tasks, exploring
the space of a combinatorial number of subtasks is extremely inefficient.

In this work, we propose to tackle the problem of hierarchical \textit{and} compositional tasks.
Prior work has tackled learning hierarchical task structures by modelling
dependencies between subtasks in a \textit{graph structure}~\cite{sohn2018hierarchical, sohn2020meta,xu2018neural,huang2019neural}.
In these settings, during training, the agent tries to efficiently
adapt to a task by inferring the latent graph structure, then uses the inferred
graph to maximize reward during test.
However, this approach does not scale for compositional tasks.
Prior work tries to infer the structure of subtasks \textit{individiually}
-- they do not consider the relations between compositional tasks.

We propose the \textit{parameterized subtask graph inference} (\psgi) approach for
tackling hierarchical and compositional tasks.
We present an overview of our approach in Figure~\ref{fig:psgi-overview}.
This approach extends the problem introduced by~\cite{sohn2020meta}.
Similar to \cite{sohn2020meta},
we assume options~\cite{sutton1999between} (low level policies) for completing subtasks have been trained
or are given as subroutines for the agent. These options are imperfect, and
require certain conditions on the state to be meet before they can be successfully executed.
We model the problem as a transfer RL problem.
During training,
an exploration policy gathers trajectories.
These trajectories are then used to infer the latent \textit{parameterized subtask graph},
$\widehat{\FG}$. $\widehat{\FG}$ models the hierarchies between compositional
tasks and options in symbolic graph structure (shown in~\ref{fig:psgi-overview}).
In \psgi, we infer the \textit{preconditions} of options, subtasks that
must be completed before an option can be successfully executed, and the \textit{effects}
of options, subtasks that are completed after they are executed.
The parameterized subtask graph is then used to maximize reward in the test environment
by using GRProp, a method introduced by~\cite{sohn2018hierarchical} which propagates
a gradient through $\widehat{\FG}$ to learn the test policy.

In \psgi, we use \textit{parameterized} options and subtasks.
This allows \psgi{} to infer the latent task structure
in a \textit{first-order logic} manner.
For example, in the cooking task in Figure~\ref{fig:psgi-overview}
we represent all \texttt{Pickup}-object options using a \textit{parameratized option},
\texttt{Pickup}~$x$.
Representing options and subtasks in parameterized form serves two roles:
1. The resulting graph is more \textbf{compact}.
There is less redundancy when representing compositional tasks that share common structure.
Hence a parameterized subtask graph requires less samples
to infer (e.g. relations for \texttt{Pickup apple}, \texttt{Pickup pan}, etc.\
are inferred at once with \texttt{Pickup}~$x$).
2. The resulting graph can \textbf{generalize} to \textit{unseen}
subtasks, where unseen subtasks may share similar structure but are not
encountered during adaptation (e.g. \texttt{Pickup cabbage} in Figure~\ref{fig:psgi-overview}).

To enable parameterized representation, we also learn the \textit{attributes}
of the components in the compositional tasks.
These attributes are used to differentiate structures of parameterized
options and subtasks. For example, in the cooking task in Figure~\ref{fig:psgi-overview},
not every object can be picked up with $\texttt{Pickup}$,
so the inferred attribute $\widehat{f}_\text{pickupable}(x)$ is a precondition
to $\texttt{Pickup}(x)$. Similarly, in a more complex cooking task,
some object $x$ may need to be sliced, before it can be boiled (e.g.\ cabbage), but some
do not (e.g.\ egg). We model these structures using \textit{parameter attributes},
$\widehat{A}_\text{att}$ (in the cooking task case objects are parameters).
We present a simple scheme to infer attributes in a \textit{zero-shot} manner,
where we infer attributes that are useful to infer relations between parameters
without supervision. These attributes are then used to generalize to
other parameters (or entities), that may be unseen during adaptation.

We summarize our work as follows:
\vspace{-4pt}
\begin{itemize}
\setlength\itemsep{-0.2em}
    \item We propose the approach of \textit{parameterized subtask graph inference} (\psgi)
    to efficiently infer the subtask structure of hierarchical and compositional
    tasks in a \textbf{first order logic manner}.
    \item We propose a simple \textit{zero-shot} learning scheme to
    infer \textit{entity attributes}, which are used to relate
    the structures of compositional subtasks.
    \item We demonstrate \psgi{} on a symbolic cooking environment that has
    complex hierarchical and compositional task structure. We show \psgi{}
    can accurately infer this structure more \textit{efficiently}
    than prior work and \textit{generalize} this structure to unseen
    tasks.
\end{itemize}
\vspace{-3pt}
\cutsectionup
\section{Problem Definition}
\cutsectiondown
\subsection{Background: Transfer Reinforcement Learning}

A \textit{task} is characterized by an MDP
$\mathcal{M}_G = \langle \mathcal{A}, \mathcal{S}, \mathcal{T}_G, \mathcal{R}_G \rangle$,
which is parameterized by a task-specific $G$, with an action space $\mathcal{A}$,
state space $\mathcal{S}$, transition dynamics $\mathcal{T}_G$, and reward function $\mathcal{R}_G$.
In the transfer RL formulation
\todo{transfer RL citations?}~\cite{duan2016rl,finn2017model}, an agent is given
a fixed distribution of training tasks $\mathcal{M}^{\text{train}}$, and must learn to
efficiently solve a distribution of unseen test tasks $\mathcal{M}^{\text{test}}$.
Although these distributions are disjoint, we assume there is some similarity
between tasks such that some learned behavior in training tasks may be useful
for learning test tasks. In each task, the agent is given $k$ timesteps to interact
with the environment (the \textit{adaptation} phase), in order to adapt to the given task.
After, the agent is evaluated on its adaptation (the \textit{test} phase).
The agent's performance is measured in terms of the expected return:
\begin{equation}
\mathcal{R}_{\mathcal{M}_G} = \mathbb{E}_{\pi_k, \mathcal{M}_G} \left[ \sum_{t=1}^H r_t \right]
\end{equation}
where $\pi_K$ is the policy after $k$ timesteps of the adaptation phase, $H$ is the
horizon in the test phase, and $r_t$ is the reward at time $t$ of the test phase.

\subsection{Background: The Subtask Graph Problem}
\label{sec:subtask-graph-problem}

The subtask graph inference problem is a transfer RL problem where tasks
are parameterized by hierarchies of \textit{subtasks}~\cite{sohn2020meta}, $\subtaskspace$.
A task is composed of $N$ subtasks, $\{ \subtask^1, \dots, \subtask^N \} \subset \subtaskspace$,
where each subtask $\subtask \in \subtaskspace$ is parameterized by the tuple
$\langle \mathcal{S}_{\text{comp}}, G_r \rangle$,
a \textit{completion set} $\mathcal{S}_{\text{comp}} \subset \mathcal{S}$,
and a \textit{subtask reward} $G_r : \mathcal{S} \to \mathbb{R}$.
The completion set $\mathcal{S}_{\text{comp}}$ denotes whether the subtask $\subtask$ is
\textit{complete}, and the subtask reward $G_r$ is the reward given to the agent when
it completes the subtask.

Following~\cite{sohn2020meta},
we assume the agent learns a set of options $\optionspace = \{\option^1, \option^2, \dots \}$ that \textit{completes} the corresponding subtasks~\cite{sutton1999between}.
These options can be learned by conditioning on subtask goal reaching reward: $r_t = \mathbb{I}(s_t \in \mathcal{S}^i_{\text{comp}})$.
Each option $\option \in \optionspace$ is parameterized by the tuple $\langle \pi, G_{\text{prec}}, G_{\text{effect}} \rangle$.
There is a trained policy $\pi$ corresponding to each $\option$.
These options may be \textit{eligible} at different precondition states $G_{\text{prec}} \subset \mathcal{S}$, where the agent must
be in certain states when executing the option, or the policy $\pi$ fails to execute
(also the \textit{initial set} of $\option$ following~\cite{sutton1999between}).
However, unlike~\cite{sohn2020meta}, these options may complete an unknown number of subtasks
(and even \textit{remove} subtask completion). This is parameterized by $G_{\text{effect}} \subset \mathcal{S}$ (also the \textit{termination set} of $\option$ following~\cite{sutton1999between}).
\todo{Previous work assumes singleton effect so $G_\text{effect}$ does not need to be inferred.}
\todo{cite http://people.csail.mit.edu/lpk/papers/hpn2.pdf to show precondition and effect has been used before}

\paragraph{Environment:} We assume that the subtask completion and
option eligibility is known to the agent.
(But the precondition, effect, and reward is hidden and must be inferred).
In each timestep $t$ the agent is the state $s_t = \{x_t, e_t, \text{step}_t, \text{step}_{\text{phase},t}, \text{obs}_t\}$.
\begin{itemize}
    \item \textbf{Completion:} $x_t \in \{0, 1\}^N$ denotes which subtasks are complete.
    \item \textbf{Eligibility:} $e_t \in \{0, 1\}^M$ denotes which options are eligible.
    \item \textbf{Time Budget:} $\text{step}_t \in \mathbb{Z}_{>0}$ is the number steps remaining in the episode.
    \item \textbf{Adaptation Budget:} $\text{step}_{\text{phase},t} \in \mathbb{Z}_{>0}$ is the number steps remaining in the adaptation phase.
    \item \textbf{Observation:} $\text{obs}_{t} \in \mathbb{R}^d$ is a low level observation of the environment at time t.
\end{itemize}

\subsection{The Parameterized Subtask Graph Problem}
\label{sec:parameterized-subtask-graph-problem}

\paragraph{Subtasks and Option Entities}
In the real world, compositional subtasks can be described in terms of a set of \textit{entities}, $\entityspace$.
(e.g.\ \texttt{pickup}, \texttt{apple}, \texttt{pear}, $\dots \in \entityspace$) that can be recombined to form new subtasks
(e.g.\ (\texttt{pickup}, \texttt{apple}), and (\texttt{pickup}, \texttt{pear})).
We assume that these entities are given to the agent.
Similarly, the learned options that execute these subtasks can also be parameterized
by the same entities
(e.g.\ [\texttt{pickup}, \texttt{apple}], and [\texttt{pickup}, \texttt{pear}]).

In real world tasks, we expect learned options with entities
that share ``attributes'' to have similar policy, precondition, and effect,
as they are used to execute subtasks with similar entities~\todo{cite}.
For example, options [\texttt{cook}, \texttt{egg}, \texttt{pot}]
and [\texttt{cook}, \texttt{cabbage}, \texttt{pot}]
share similar preconditions (the target ingredient must be placed in the \textit{pot}),
but also different (\textit{cabbage} must be sliced, but the egg does not).
In this example, \textit{egg} and \textit{cabbage} are both \textit{boilable}
entities, but \textit{egg} is not \textit{sliceable}.
\todo{match examples to Figure 1!}

To model these similarities, we assume in each task, there
exist boolean \textit{latent attribute functions} which indicate
shared attributes in entities. E.g. $f_\text{pickupable} : \entityspace \to \{0, 1\}$,
where $f_\text{pickupable}(\texttt{apple}) = 1$.
We will later try to infer the values of these latent entities,
so we additionally assume there exist some weak supervision, where
a low-level embedding of entities is provided to the agent,
$\entityembed : \entityspace \to \mathbb{R}^D$.

\paragraph{The Parameterized Subtask Graph}
Our goal is to infer the underlying task structure between subtasks and options
so that the agent may complete subtasks in an optimal order.
As defined in the previous sections, this task structure can be completely determined
by the option preconditions, option effects, and subtask rewards.
As such we define the \textit{parameterized subtask graph} to be the tuple of the \textit{parameterized} preconditions, effects, and rewards for all subtasks and options:
\begin{equation}
    \FG = \langle \FGpcond, \FGeffect, \FGr \rangle
\end{equation}
where
$\FGpcond : \entityspace^N \times \mathcal{S} \to \{0, 1\},
\FGeffect : \entityspace^N \times \mathcal{S} \to \mathcal{S},
$ and $\FGr : \entityspace^N \times \mathcal{S} \to \mathbb{R}$.
The parameterized precondition, $\FGpcond$, is a function
from an option with $N$ entities and a subtask completion set to $\{0, 1\}$, which
specifies whether the option is eligible under a completion set.
E.g. If $\FGpcond([X_1, X_2], s) = 1$, then
the option $[X_1, X_2]$ is eligible if the agent is in state $s$.
The parameterized effect, $\FGeffect$, is a function from an option with $N$ entities
and subtask completion set to a different completion set.
Finally, the parameterized reward, $\FGr$, is a function from a subtask with $N$ entities
to the reward given to the agent from executing that subtask.

Our previous assumption that options with similar entities and attributes
share preconditions and effects manifests in $\FGpcond$ and $\FGeffect$
where these functions tend to be \textit{smooth}.
Similar inputs to the function (similar option entities) tend to yield
similar output (similar eligibility and effect values).
This smoothness gives two benefits.
1. We can share experience between similar options for inferring preconditions and effect.
2. This enables generalization to preconditions and effects of unseen entities.
Note that this smoothness does \textit{not} apply to the reward $\FGr$.
We assume reward given for subtask completion is independent across tasks.

\begin{figure*}[h!]
  \centering
  \includegraphics[width=0.95\textwidth]{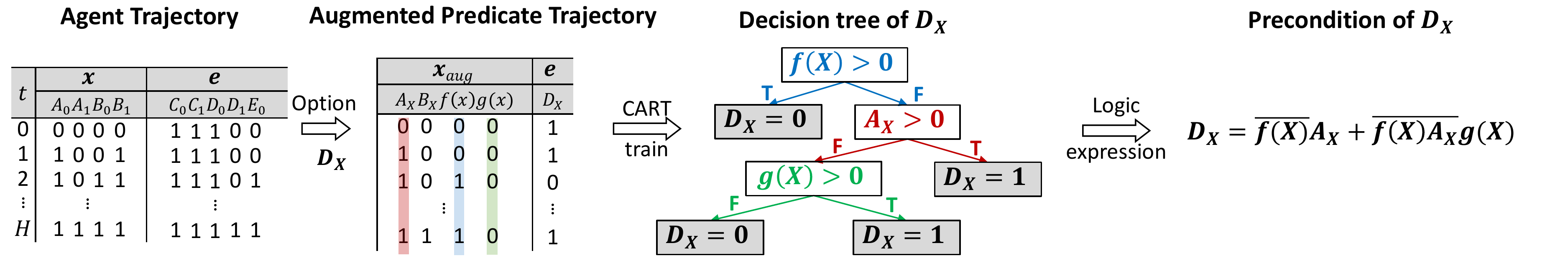}
  \vspace{-10pt}
  \caption{An overview of our approach for estimating the parameterized precondition of the subtask graph $\widehat{\FGpcond}$ in a simple environment with subtasks $A, B$ and options $C, D, E$.
  Each subtask and option has a parameter 0 or 1. Note by inferring the parameterized
  precondition and effects, we can infer the behavior unseen subtasks and options such as $D_2$.
  We run precondition inference for every option and show $D_X$ as an example.
  \textbf{1.} The first table is built from the agent's trajectory ($x$ is the subtask completion, $e$ the option eligibility).
  \textbf{2.} We build the second table, the ``augmented'' trajectory by substituting
  $X$ into all possible subtask completions, $A_X, B_X$, and inferred attributes $f, g$.
  \textbf{3.} We train a decision tree over the table, to infer the relation $x_\text{aug} \to D_x$ (predicting when $D_x$ is eligible given the completion $x_\text{aug}$).
  \textbf{4.} We translate the decision tree into an equivalent predicate boolean expression,
  which is one part of the inferred parameratized subtask graph \smash{$\widehat{\FG}$}.
  }
  \label{fig:pilp-overview}
  \vspace{-10pt}
\end{figure*}

\vspace{-5pt}
\cutsectionup
\section{Method}
\cutsectiondown
\label{sec:method}

We propose the \textit{Parameterized Subtask Graph Inference} (\psgi) method to efficiently infer the latent
parameterized subtask graph $\FG = \langle \FGpcond, \FGeffect, \FGr \rangle$.
Figure~\ref{fig:pilp-overview} gives an overview of our approach.
At a high level, we use the adaptation phase to gather \textit{adaptation trajectories}
from the environment using an adaptation policy $\pi_\theta^{\text{adapt}}$.
Then, we use the adaptation trajectories to infer the latent subtask graph $\widehat{\FG}$.
In the test phase, a \textit{test policy} $\pi_{\widehat{\FG}}^{\text{test}}$ is conditioned
on the inferred subtask graph $\widehat{\FG}$ and maximizes the reward.
As the performance of the test policy is dependent on the inferred subtask graph $\widehat{\FG}$, it is important to accurately infer this graph.
Note that the test task may contain subtasks that are \textit{unseen} in the training task.
We learn a predicate subtask graph $\widehat{\FG}$ that can \textit{generalize} to
these unseen subtasks and options.

\subsection{Zero-shot Learning Entity Attributes}
\label{sec:zero-shot-learning-attributes}
In the \textit{Parameterized Subtask Graph Problem} definition,
we assume there exist \textit{latent attributes} that indicate
shared structure between options and subtasks with the same attributes.
E.g. One attribute may be $f_\text{pickupable} : \entityspace \to \{0, 1\}$,
where $f_\text{pickupable}(\texttt{apple}) = 1$, etc.
Our goal is to infer a set of candidate attribute functions,
$\widehat{A}_\text{att} = \{\widehat{f}_1, \widehat{f}_2, \dots \}$, such that options
with the same attributes indicates the same preconditions.
As there is no supervision involved,
we formulate this inference as a \textit{zero shot learning problem}~\cite{Palatucci2009ZeroshotLW}.
Note the inferred attributes that are preconditions for options should not only
construct an accurate predicate subtask graph for options seen in the adaptation
phase, but also unseen options.

During the adaptation phase, the agent will encounter a set of seen
entities $E \subset \entityspace$. We construct candidate attributes
from $E$ using our \textit{smoothness} assumption,
where similar entities result in similar preconditions. We generate
candidate attributes based on similarity using the given entity embedding,
$\entityembed : \entityspace \to \mathbb{R}^D$.

Let $C = \{C_1, C_2, \dots \}$ be an exhaustive set of clusters generated
from $E$ using $\entityembed$. Then, we define a candidate attribute
function from each cluster:
$\widehat{f}_i(X) := \mathbb{I}[X \in C_i]$
To infer the attribute of an unseen entity $X \not \in E$, we use a 1-Nearest Neighbor
classifier that uses the attributes of the nearest seen entity~\cite{fix1985discriminatory}.
$\widehat{f}_i(X) = \mathbb{I}[X^* \in C_i]$
where
$X^* = \argmin_{X' \in E} \text{dist}(\entityembed(X), \entityembed(X'))$.

\subsection{Parameterized Subtask Graph Inference}

Let
$\tau_H = \{ s_1, o_1, r_1, d_1, \dots, s_H\}$ be the adaptation trajectory
of the adaptation policy $\pi_\theta^{\text{adapt}}$ after $H$ time steps.
Our goal is to infer the maximum likelihood parameterized subtask graph
$\FG$ given this trajectory $\tau_H$.
\begin{equation}
    \widehat{\FG}^\text{MLE} = \argmax_{\FGpcond, \FGeffect, \FGr} p(\tau_H | \FGpcond, \FGeffect, \FGr)
\end{equation}

By expanding this likelihood term, we show that to maximize $\widehat{\FG}$,
it suffices to maximize $\hatFGpcond$, $\hatFGeffect$, and $\hatFGr$ individually.
\begin{align}
    \widehat{\FG}^\text{MLE}
    =& \left(\hatFGpcond^\text{MLE}, \hatFGeffect^\text{MLE}, \hatFGr^\text{MLE}\right)
    \\=& \bigg(
    \argmax_{\FGpcond}\prod_{t=1}^H p(e_t | x_t, \FGpcond),
    \\&\argmax_{\FGeffect}\prod_{t=1}^H p(x_{t+1} | x_t, o_t, \FGeffect),\label{eq:mle-exp-2}
    \\&\argmax_{\FGr}\prod_{t=1}^H p(r_t | o_t, o_{t+1}, \FGr)\label{eq:mle-exp-3}
    \bigg)
\end{align}
We show details of this derivation in the appendix.
Next, we explain how to compute $\hatFGpcond$, $\hatFGeffect$, and $\hatFGr$.

\paragraph{Parameterized Precondition Inference via Predicate Logic Induction}
We give an overview of how we infer the option preconditions $\hatFGpcond$ in Figure~\ref{fig:pilp-overview}.
Note from the definition,
we can view the precondition $\FGpcond$ as a deterministic function,
$f_{\FGpcond} : (E, x) \mapsto \{0, 1\}$,
where $E$ is the option entities, and $x$ is the completion set vector.
Hence, the probability term in Eq.(\ref{eq:mle-exp-2}) can be written as
$p(e_t | x_t, \FGpcond) =
\prod_{i=1}^N \mathbb{I}[e_t^{(i)} = f_{\FGpcond}(E^{(i)}, x_t)]$
where $\mathbb{I}$ is the indicator function, and $E^{(i)}$ is the entity
set of the $i$th option in the given task. Thus, we have
\begin{equation}
    \hatFGpcond^\text{MLE} = \argmax_{\FGpcond} \prod_{t=1}^H \prod_{i=1}^N \mathbb{I}[e_t^{(i)} = f_{\FGpcond}(E^{(i)}, x_t)]
    \label{eq:mle-prec}
\end{equation}

Following~\cite{sohn2020meta}, this can be maximized by finding
a boolean function $\widehat{f}_{\FGpcond}$ over \textit{only} subtask completions $x_t$ that satisfies all the indicator functions in Eq.(\ref{eq:mle-prec}).
However this yields multiple possible solutions --- particularly the preconditions
of unseen option entities in the trajectory $\tau_H$.
If we infer a $\widehat{f}_{\FGpcond}$ separately over all seen options (without considering
the option parameters), this solution is identical to the solution proposed by~\cite{sohn2020meta}.
We want to additionally generalize our solution over multiple unseen subtasks and options
using the entities, $E$.

We leverage our smoothness assumption --- that 
$\widehat{f}_{\FGpcond}$ is \textit{smooth} with respect to the input entities and attributes.
E.g.\ If the inferred precondition for the option [\texttt{pickup}, $X$] is
the candidate attribute $\widehat{f}(X)$, any entity $X$ where $\widehat{f}(X) = 1$
has the same precondition.
I.e. For some unseen entity set $E^*$ we want the following property to hold:\
\begin{equation}
\widehat{f}_i(E) = \widehat{f}_i(E^*) \; \text{for some $i$} \; \Rightarrow
\widehat{f_{\FGpcond}}(E, x_t) = \widehat{f_{\FGpcond}}(E^*, x_t)
\end{equation}

To do this, we infer a boolean function $\widehat{f_{\FGpcond}}$ over
\textit{both} subtask completions $x_t$ \textit{and} entity variables $X \in E$.
We use (previously inferred) candidate attributes over entities, $\widehat{f}_i(X) \forall X \in E$ in the boolean function to serve as \textit{quantifiers}.
Inferring in this manner insures that the precondition function 
$\widehat{f}_{\FGpcond}$ is \textit{smooth} with respect to the input entities and attributes.
Note that some but not all attributes may be shared in entities.
E.g. [\texttt{cook}, \texttt{cabbage}] has similar but not the same
preconditions as [\texttt{cook}, \texttt{egg}].
So, we cannot directly reuse the same preconditions for similar entities.
We want to generalize between different \textit{combinations} of attributes.

We translate this problem as an \textit{inductive logic programming} (ILP)
problem~\cite{muggleton1994inductive}.
We infer the eligibility (boolean output) of some option $\mathcal{O}$ with some
entities(s) $E = \{X_1, X_2, \dots\}$, from boolean input formed by all possible
completion values $\{x^i_t\}^H_{t=1}$,
and all attribute values $\{\widehat{f}_i(X)\}^{i = 1\dots }_{X \in E}$.
We use the \textit{classification and regression tree} (CART) with Gini impurity
to infer the the precondition functions $\widehat{f}_{\FGpcond}$ for each parameter $E$ \cite{breiman1984classification}.
Finally, the inferred decision tree is converted into an equivalent symbolic logic
expression and used to build the parameterized subtask graph.

\paragraph{Parameterized Effect Inference}
We include an visualization of how we infer the option effects $\hatFGeffect$ in the appendix
in the interest of space.
From the definitions of the parameterized subtask graph problem, we can write the predicate option effect $\FGeffect$
as a deterministic function $f_{\FGeffect} : (E, x_t) \mapsto x_{t+1}$,
where if there is subtask completion $x_t$, executing option $\mathcal{O}$ (with entities
$E$) successfully results in subtask completion $x_{t+1}$.
Similar to precondition inference, we have
\begin{equation}
    \hatFGeffect^\text{MLE} = \argmax_{\FGeffect}
    \prod_{t=1}^H \prod_{i=1}^N \mathbb{I}[x_{t+1} = f_{\FGeffect}(E^{(i)}, x_t)]
\end{equation}
As this is deterministic, we can calculate the element-wise difference between $x_t$ (before option) and
$x_{t+1}$ (after option) to infer $f_{\FGeffect}$.
\begin{equation}
    \widehat{f_{\FGeffect}}(E^{(i)}, x) = x + \mathbb{E}_{t=1\dots H}[x_{t+1} - x_t | o_t = \mathcal{O}^{(i)}]
    \label{eq:non-smooth-effect}
\end{equation}

Similar to precondition inference, we also want to infer the effect of options with
unseen parameters. We leverage the same smoothness assumption:
\begin{equation}
\widehat{f}_i(E) = \widehat{f}_i(E^*) \; \text{for some $i$} \; \Rightarrow
\widehat{f_{\FGeffect}}(E, x_t) = \widehat{f_{\FGeffect}}(E^*, x_t)
\label{eq:effect-smooth-assumption}
\end{equation}
Unlike preconditions, we expect the effect function to be relatively constant
across attributes, i.e., the effect of executing option [\texttt{cook}, $X$]
is always completing the subtask (\texttt{cooked}, $X$), no matter the attributes
of $X$.
So we directly set the effect of unseen entities, $\widehat{f_{\FGeffect}}(E^*, x_t)$, by similarity according to
Equation~\ref{eq:effect-smooth-assumption}.

\paragraph{Reward Inference}
We model the subtask reward as a Gaussian distribution
$\FGr(E) \sim \mathcal{N}(\widehat{\mu}_{E}, \widehat{\sigma}_{E})$.
The MLE estimate of the subtask reward becomes the empirical mean
of the rewards received during the adaptation phase when subtask with parameter $\mathcal{T}$
becomes complete. For the $i$th subtask in the task with entities $E^i$,
\begin{equation}
    \hatFGr(E^i) = \widehat{\mu}_{E^i}
    = \mathbb{E}_{t = 1\dots N}[r_t | x^i_{t+1} - x^i_t = 1]
    \label{eq:reward-estimation}
\end{equation}
Note we do not use the smoothness assumption for $\hatFGr(E)$,
as we assume reward is independently distributed across tasks. We
initialize $\hatFGr(E^*) = 0$ for unseen subtasks with entities $E^*$
and update these estimates with further observation.

\subsection{Task Transfer and Adaptation}
In the test phase, we instantiate a \textit{test policy}
$\pi_{\widehat{\FG}_\text{prior}}^\text{test}$ using the  parameterized 
subtask graph \smash{$\widehat{\FG}_\text{prior}$}, inferred from the training task samples.
The goal of the test policy is to maximize reward in the test environment using
$\widehat{\FG}_\text{prior}$.
As we assume the reward is independent across tasks, we re-estimate the reward
of the test task according to Equation~\ref{eq:reward-estimation}, without
task transfer.
With the reward inferred, this yields the same problem setting
given in~\cite{sohn2018hierarchical}.
\cite{sohn2018hierarchical} tackle this problem using GRProp, which models the subtask
graph as differentiable function over reward, so that the test policy
has a dense signal on which options to execute are likely to maximally increase the reward.

However, the inferred parameterized subtask graph may be imperfect,
the inferred precondition and effects may not transfer to the test task.
To adapt to possibly new preconditions and effects, we use samples gathered
in the adaptation phase of the test task to infer a new parameterized subtask
graph \smash{$\widehat{\FG}_\text{test}$},
which we use to similarly instantiate another test policy
$\pi_{\widehat{\FG}_\text{test}}^\text{test}$ using GRProp.
We expect \smash{$\widehat{\FG}_\text{test}$} to eventually be more accurate
than \smash{$\widehat{\FG}_\text{prior}$} as more timesteps are gathered in the test environment.
%
To maximize performance on test, we thus choose to instantiate
a posterior test policy $\pi^\text{test}_\text{posterior}$, which is
an \textit{ensemble} policy over $\pi^\text{test}_{\widehat{\FG}_\text{prior}}$
and $\pi^\text{test}_{\widehat{\FG}_\text{test}}$.
We heuristically set the weights of $\pi^\text{test}_\text{posterior}$
to favor $\pi^\text{test}_{\widehat{\FG}_\text{prior}}$
early in the test phase, and 
$\pi^\text{test}_{\widehat{\FG}_\text{test}}$
later in the test phase.
\cutsectionup
\section{Related Work}
\cutsectiondown

\textbf{Subtask Graph Inference.} The subtask graph inference (SGI) framework~\citep{sohn2018hierarchical,sohn2020meta} assumes that a task consists of multiple base subtasks, such that the entire task can be solved by completing a set of subtasks in the right order. Then, it has been shown that SGI can efficiently solve the complex task by explicitly inferring the precondition relationship between subtasks in the form of a graph using an inductive logic programming (ILP) method. The inferred subtask graph is in turn fed to an execution policy that can predict the optimal sequence of subtasks to be completed to solve the given task.

However, the proposed SGI framework is limited to a single task; the knowledge learned in one task cannot be transferred to another. This limits the SGI framework such that
does not scale well to compositional tasks, and cannot generalize to unseen tasks.
We extend the SGI framework by modeling \textit{parameterized} subtasks and options,
which encode relations between tasks to allow efficient and general learning.
In addition, we generalize the SGI framework by learning an effect model --
In the SGI framework it was assumed that for each subtask there is a corresponding option, that
completes that subtask (and does not effect any other subtask).
%
\cutparagraphup
\textbf{Compositional Task Generalization.}
Prior work has also tackled compositional generalization in a symbolic manner~\citep{loula2018rearranging,andreas2017modular,oh2017zero}.
\citet{loula2018rearranging} test compositional generalization of
natural language sentences in recurrent neural networks.
\citet{andreas2017modular,oh2017zero} tackle compositional task generalization
in an \textit{instruction following} context, where an agent
is given a natural language instruction describing the task the agent must complete
(e.g.\ ``pickup apple").
These works use \textit{analogy making} to learn policies that can execute
instructions by analogy (e.g.\ ``pickup $X$").
However, these works construct policies on the \textit{option level} --
they construct policies that can execute ``pickup $X$" on different $X$ values.
They also do not consider hierarchical structure for the order which options should
be executed (as the option order is given in instruction).
Our work aims to learn these analogy-like relations at a between-options level,
where certain subtasks must be completed before another option can be executed.

\cutparagraphup
\textbf{Classical Planning.}
At a high level, a parameterized subtask graph $\FG$ is similar to a STRIPS planning domain~\citep{fikes1971strips} with an attribute model add-on~\citep{frank2003constraint}.
Prior work in classical planning has proposed to learn STRIPS domain specifications (action schemas) through given trajectories (action traces)~\citep{suarez2020strips,mehta2011autonomous,walsh2008efficient,zhuo2010learning}.
Our work differs from these in 3 major ways:
1. \psgi{} learns an \textit{attribute} model, which is crucial to generalizing
compositional tasks with components of different behaviors.
2. We evaluate \psgi{} on more hierarchical domains, where prior work
has evaluated on pickup-place/travelling classical planning problems, which admit
flat structure.
3. We evaluate \psgi{} on generalization, where there may exist subtasks and options
that are not seen during adaptation.
\cutsectionup
\section{Experiments}
\cutsectiondown

We aim to answer the following questions:
\begin{enumerate}
    \item Can PSGI \textit{generalize} to unseen evaluation tasks in zero-shot manner by transferring the inferred task structure?
    \item Does PSGI \textit{efficiently} infers the latent task structure compared to prior work (MSGI~\citep{sohn2020meta})?
\end{enumerate}


\begin{figure*}[!h]
  \centering
  \includegraphics[trim=15 35 10 38,clip,width=0.8\textwidth]
  {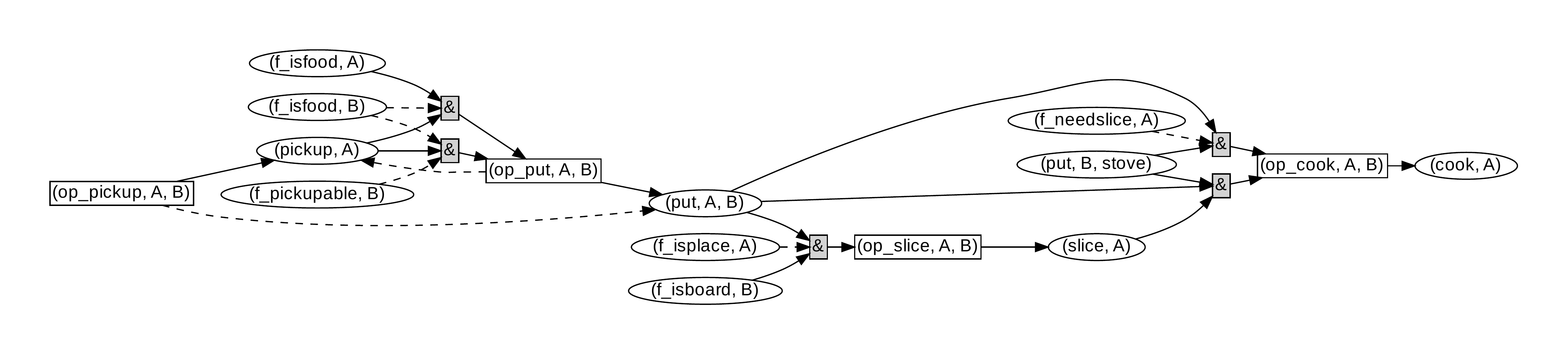}
  \vspace{-5pt}
  \caption{The inferred parameterized subtask graph by \fsgibf after 2000 timesteps in the \cook.
  Options are represented in rectangular nodes. Subtask completions and attributes are in oval nodes. A solid line represents a positive precondition / effect, dashed for negative.
  Ground truth attributes are included option/subtask parameters, however
  which attributes are used for which option preconditions is still hidden, which
  PSGI must infer.
  }
  \vspace{-5pt}
  \label{fig:cooking-pilp}
\end{figure*}
\begin{figure*}[!h]
  \centering
  \includegraphics[trim=15 10 10 50,clip,width=0.25\textwidth]
  {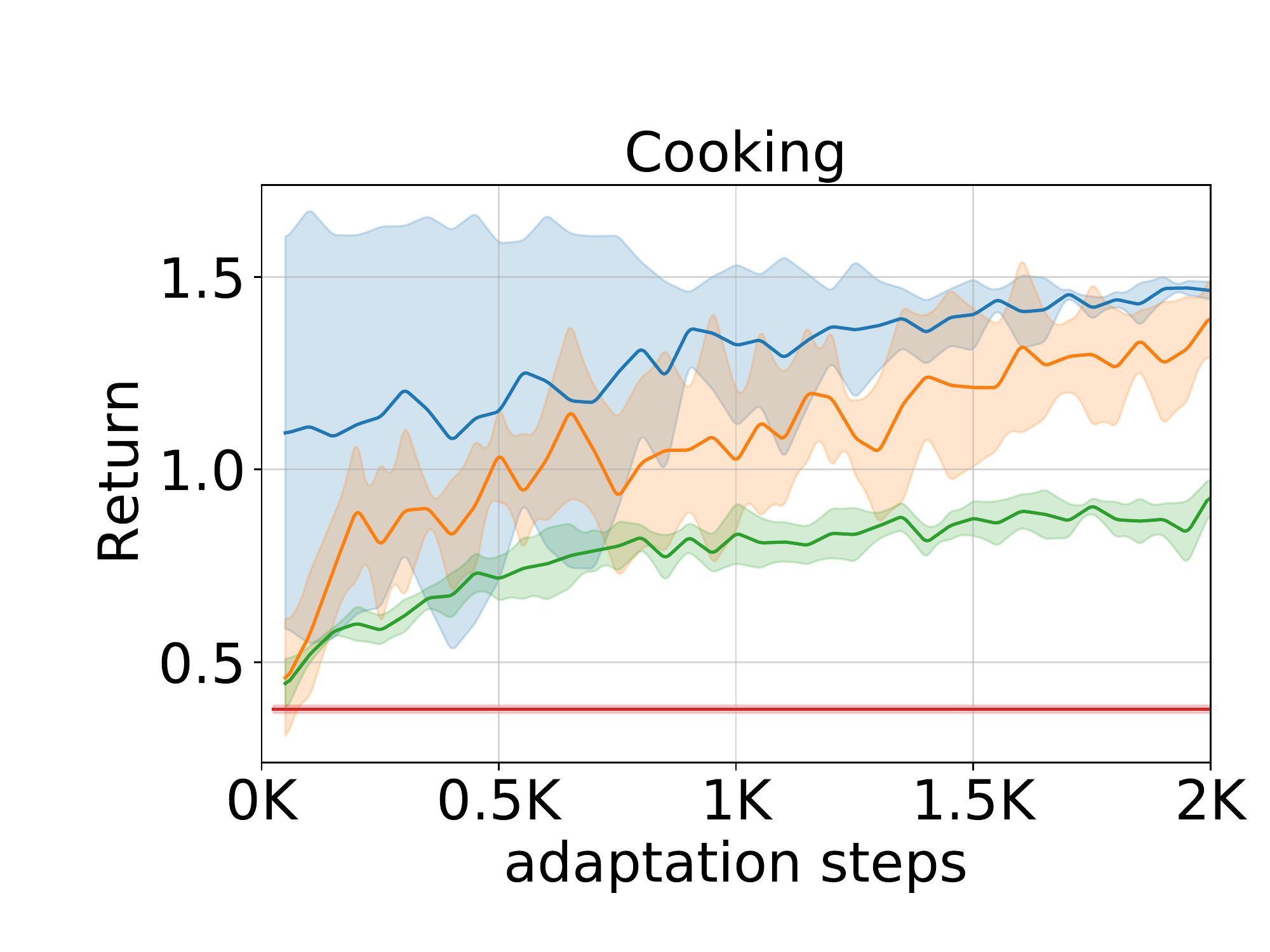}
  \includegraphics[trim=15 10 10 50,clip,width=0.25\textwidth]
  {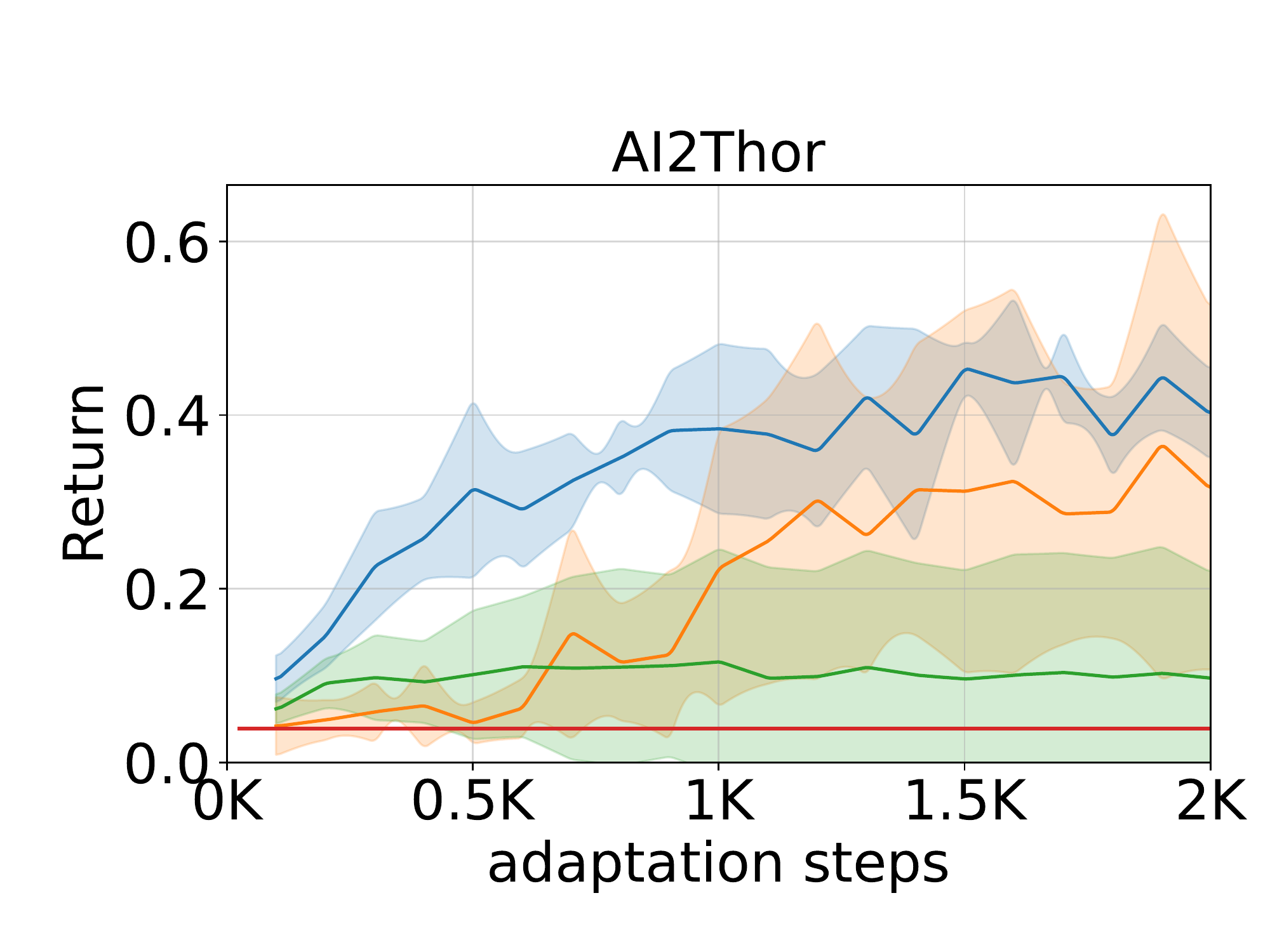}
  \includegraphics[trim=15 10 10 50,clip,width=0.25\textwidth]
  {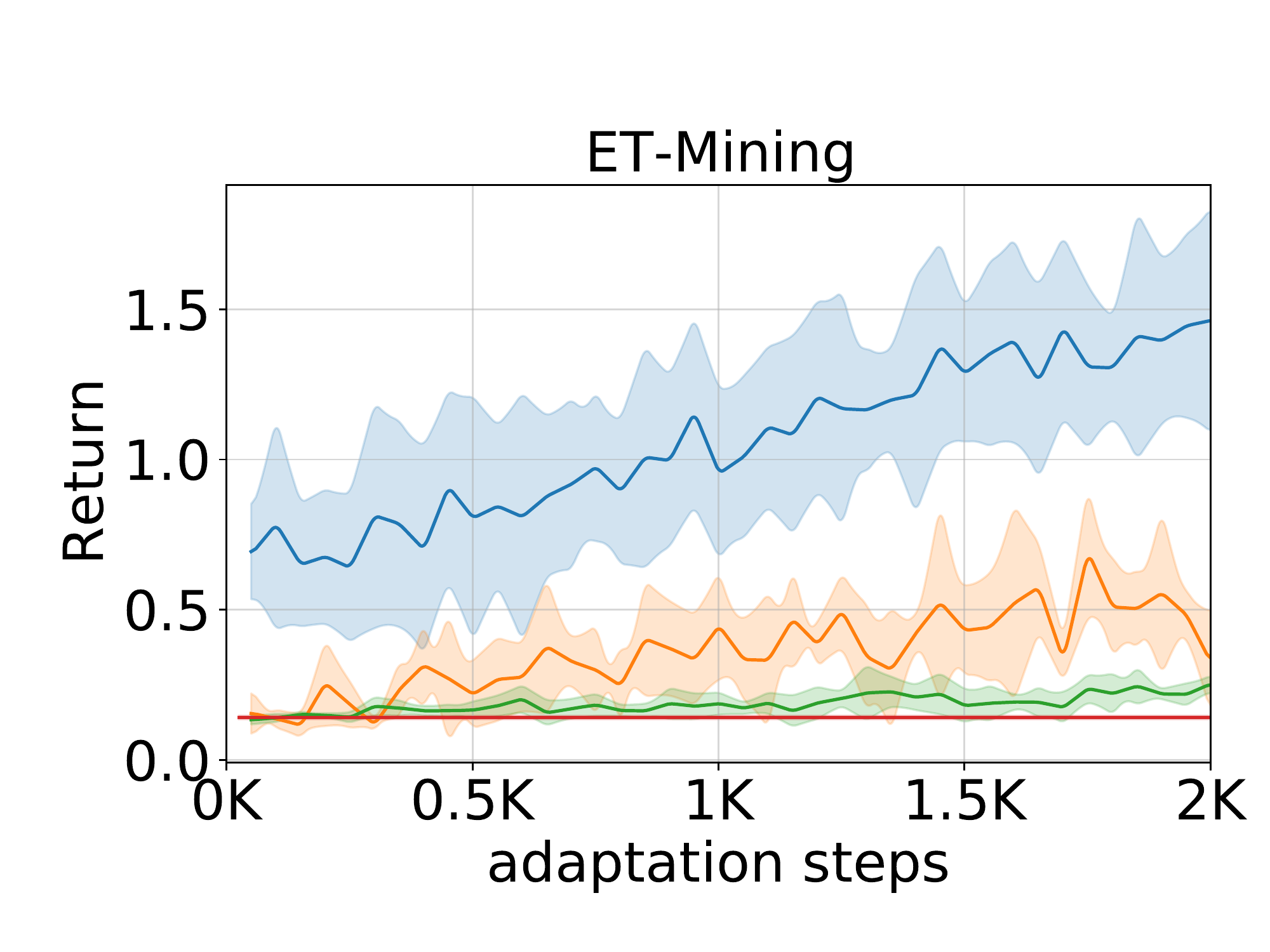}
  \\
  \includegraphics[trim=15 10 10 50,clip,width=0.25\textwidth]
  {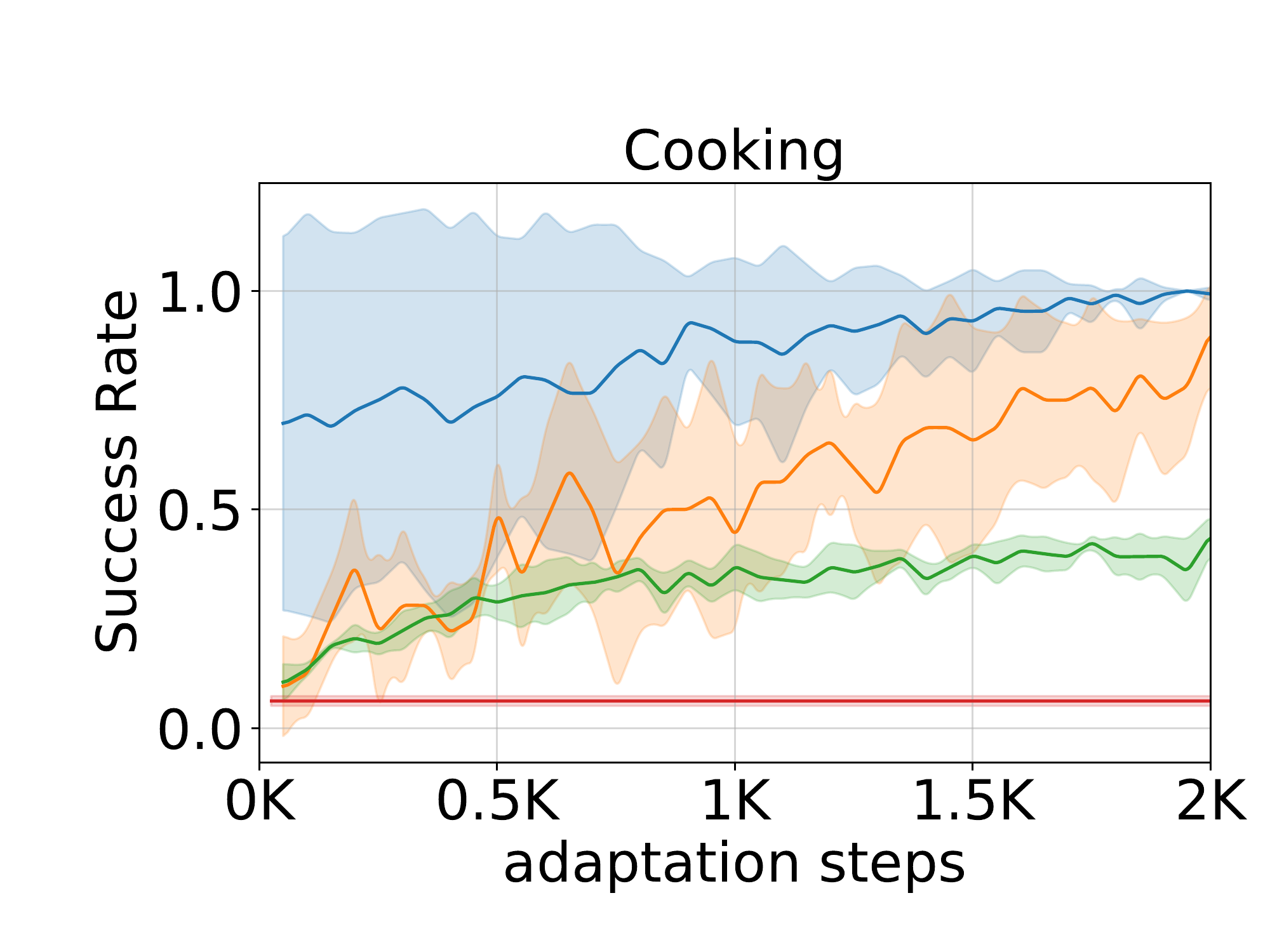}
  \includegraphics[trim=15 10 10 50,clip,width=0.25\textwidth]
  {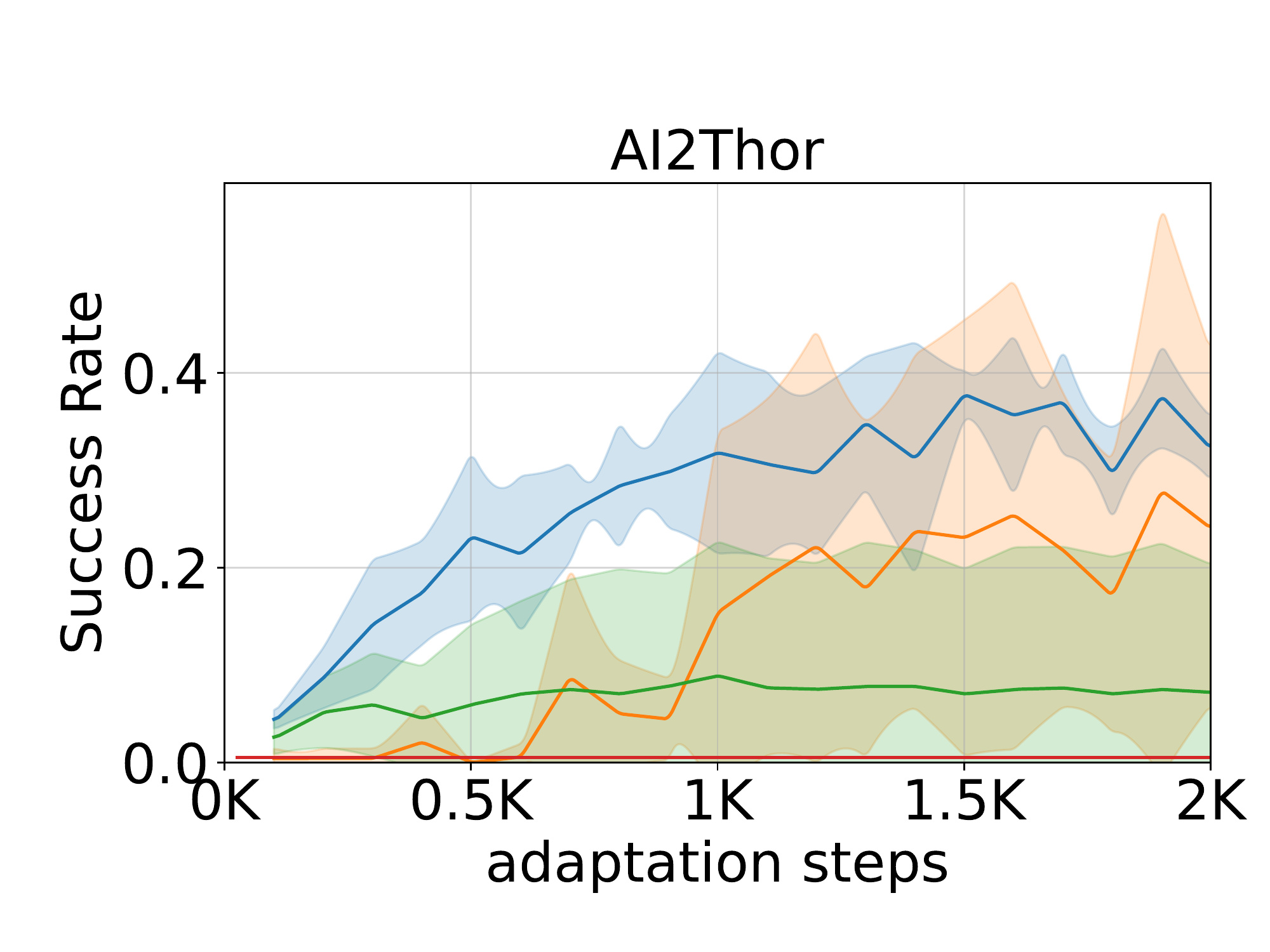}
  \includegraphics[trim=0 10 10 50,clip,width=0.25\textwidth]
  {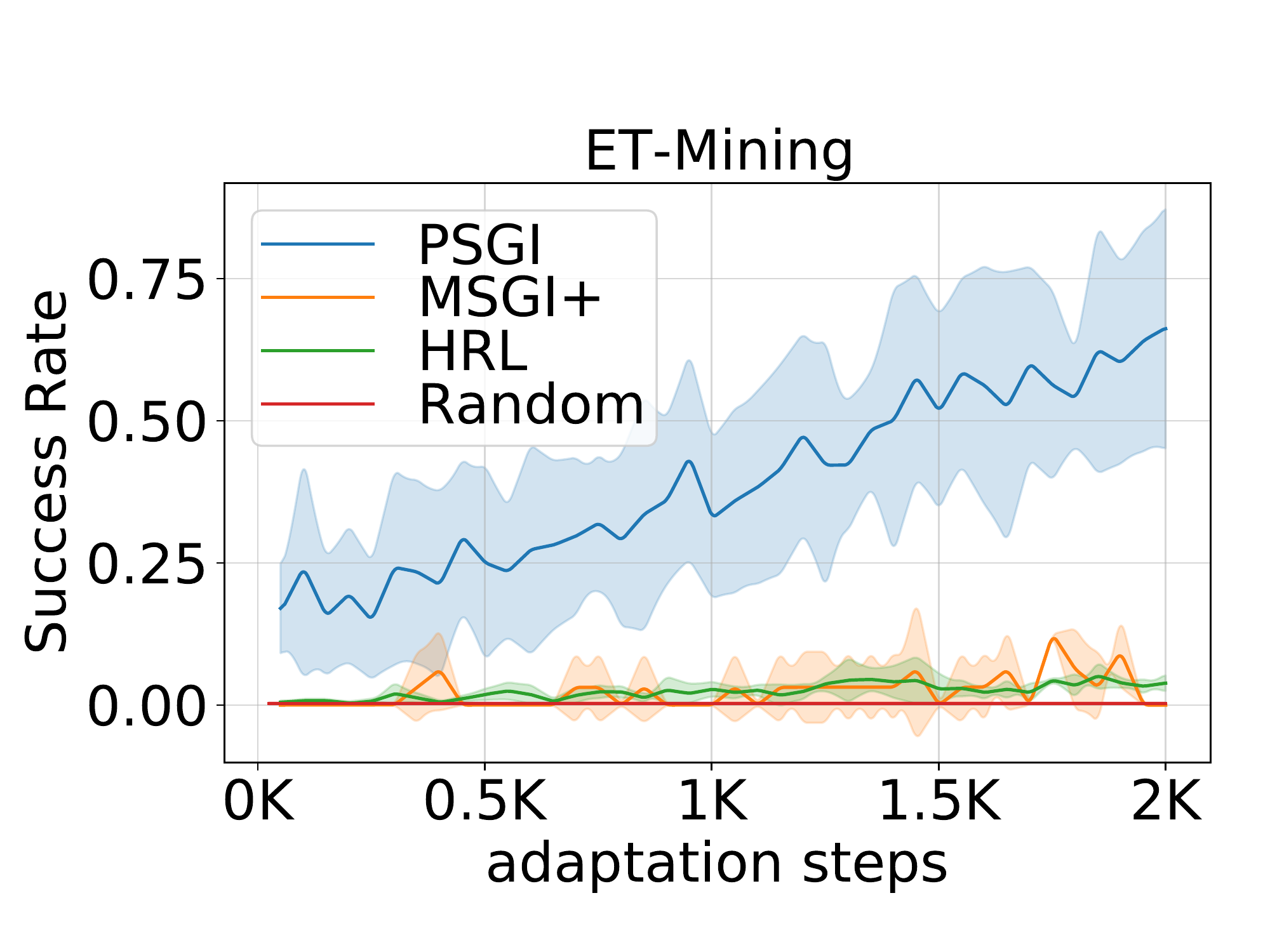}
  \vspace{-5pt}
  \caption{The adaptation curves in the \cook, \thor, and \mine domains.
  \vspace{-15pt}
  }
  \label{fig:transfer_performance}
\end{figure*}

\cutsubsectionup
\subsection{Environments}
\cutsubsectiondown

We evaluate PSGI in novel symbolic environments, \thor, \cook, and \mine.
\thor is a symbolic environment based on \citep{ai2thor}, a simulated realistic indoor environment.
In our \thor environment, the agent is given a set of pre-trained
options and must cook various food objects in different kitchen layouts,
each containing possibly unseen objects.
\cook is a simplified cooking environment with similar but simpler dynamics
to \thor. An example of the simplified \cook task is shown in Figure~\ref{fig:psgi-overview}.
The \mine domain is modelled after the open world video game Minecraft
and the domain introduced by~\citet{sohn2018hierarchical}.

\textbf{Tasks.}
In \thor, there are 30 different tasks based on the 30 kitchen floorplans in \cite{ai2thor}.
In each task, 14 entities from the floorplan are sampled at random.
Then, the subtasks and options are populated by replacing the parameters in parameterized subtasks and options by the sampled entities; \eg, we replace \texttt{X} and \texttt{Y} in the parameterized subtask \texttt{(pickup, X, Y)} by $\{\texttt{apple, cabbage, table}\}$ to populate nine subtasks. 
This results in 1764 options and 526 subtasks. The ground-truth attributes are taken from \citep{ai2thor} but are not available to the agent.
\cook is defined similarly and has a pool of 22 entities and 10 entities are chosen at random for each task.
This results in 324 options and 108 subtasks.
Similarly for \mine, we randomly sample 12 entities from a pool of 18 entities and populate 180 subtasks and 180 options for each task.
In each environment, the reward is assigned at random to one of the subtasks that have the largest critical path length, where the critical path length is the minimum number of options to be executed to complete each subtask. See the appendix for more details on the tasks.

\textbf{Observations.}
At each time step, the agent observes the completion and eligibility vectors (see section~\ref{sec:parameterized-subtask-graph-problem} for definitions) and the corresponding embeddings. The subtask and option embeddings are the concatenated vector of the embeddings of its entities; \eg, for \texttt{pickup, apple, table} the embedding is $[f(\texttt{pickup}), f(\texttt{apple}), f(\texttt{table})]$ where $f(\cdot)$ can be an image or language embeddings. In our experiments, we used 50 dimensional GloVE word embeddings~\citep{pennington2014glove} as the embedding function $f(\cdot)$.

\cutsubsectionup
\subsection{Baselines}
\cutsubsectiondown

\begin{itemize}
   \setlength{\itemsep}{0pt}
    \item \msgi is the \msgiorg~\citep{sohn2020meta} agent modified to be capable of solving our \cook and \mine tasks. We augmented \msgiorg with an effect model, separate subtasks and options in the ILP algorithm, and replace the GRProp with cyclic GRProp, a modified version of GRProp that can run with cycles in the subtask graph.
    \item \hrl~\citep{andreas2017modular}\footnote{In~\citet{andreas2017modular} this agent was referred as Independent model.} is the option-based hierarchical reinforcement learning agent. It is an actor-critic model over the pre-learned options.
    \item \random agent randomly executes any eligible option.
\end{itemize}
We meta-train \fsgibf on training tasks and meta-eval on evaluation tasks to test its adaptation efficiency and generalization ability.
We train \hrl on evaluation tasks to test its adaptation (\ie, learning) efficiency.
We evaluate \random baseline on evaluation tasks to get a reference performance.
We use the same recurrent neural network with self-attention-mechanism so that the agent can handle varying number of (unseen) parameterized subtasks and options depending on the tasks. See the appendix for more details on the baselines.

\subsection{Zero-shot Transfer Learning Performance}
\Cref{fig:transfer_performance} compares the zero-shot and few-shot transfer learning performance on \cook, \thor, and \mine domains.
First, \fsgibf achieves over 50 and 20\% success rate on \cook and \mine domain without observing any samples (\ie, x-axis value = 0) in unseen evaluation tasks. This indicates that the parameterized subtask graph effectively captures the shared task structure, and the inferred attributes generalizes well to unseen entities in zero-shot manner. Note that \msgi, \hrl, and \random baselines have no ability to transfer its policy from training tasks to unseen evaluation tasks.
\fsgibf achieves over 5\% zero shot success rate on \thor, still performing better than baselines, but relatively low compared to \fsgibf on \cook and \mine, which indicates the
high difficulty of transfer in \thor.

\subsection{Few-shot Transfer Learning Performance}
In~\Cref{fig:transfer_performance}, \fsgibf achieves over 90, 30, 80\% success rate on \cook, \thor, and \mine domains respectively after only 1000 steps of adaptation, while other baselines do not learn any meaningful policy except \msgi in \cook and \thor. This demonstrates that the parameterized subtask graph enables \fsgibf to share the experience of similar subtasks and options (\eg, \texttt{pickup X on Y} for all possible pairs of \texttt{X} and \texttt{Y}) such that the sample efficiency is increased by roughly the factor of number of entities compared to using subtask graph in \msgi.

\subsection{Comparison on Task Structure Inference}
We ran \fsgibf and \msgi in \cook, \thor, and \mine, inferring the latent
subtask graphs 
for 2000 timesteps. The visualized inferred graphs at 2000 timesteps are shown in~\Cref{fig:cooking-pilp}.
In the interest of space, we have shown the graph by \msgi in the appendix in~\Cref{fig:cooking-pilp-msgi}.
\fsgibf infers the parameterized graph using first-order logic, and thus it is more compact.
However, \msgi infers the subtask graph without parameterizing out the shared structure, resulting in a non-compact graph with hundreds of subtasks and options. Moreover, graph inferred by \fsgibf has 0\% error in precondition and effect model inference. The graph inferred by \msgi has 38\% error in the preconditions (the six options that \msgi completely failed to infer any precondition are not shown in the figure for readability).
\cutsectionup
\section{Conclusion}
\cutsectiondown

In this work we presented \textit{parameterized subtask graph inference} (\psgi),
a method for efficiently inferring the latent structure of
hierarchical and compositional tasks.
\psgi{} also facilitates inference of \textit{unseen} subtasks
during adaptation, by inferring relations using predicates.
\psgi{} additionally learns \textit{parameter attributes} in a zero-shot manner, which
differentiate the structures of different predicate subtasks.
Our experimental results showed that \psgi{} is more efficient and more general
than prior work.
In future work, we aim to to tackle noisy settings, where options
and subtasks exhibit possible failures, and settings where the option
policies must also be learned.
\section{Acknowledgements}

The authors would like to thank Yiwei Yang and Wilka Carvalho for their valuable discussions. This work was supported in part by funding from LG AI Research.

\bibliography{ref.bib}

\begin{thebibliography}{26}
\providecommand{\natexlab}[1]{#1}

\bibitem[{Andreas, Klein, and Levine(2017)}]{andreas2017modular}
Andreas, J.; Klein, D.; and Levine, S. 2017.
\newblock Modular multitask reinforcement learning with policy sketches.
\newblock In \emph{International Conference on Machine Learning}, 166--175.
  PMLR.

\bibitem[{Breiman et~al.(1984)Breiman, Friedman, Stone, and
  Olshen}]{breiman1984classification}
Breiman, L.; Friedman, J.; Stone, C.~J.; and Olshen, R.~A. 1984.
\newblock \emph{Classification and regression trees}.
\newblock CRC press.

\bibitem[{Carvalho et~al.(2020)Carvalho, Liang, Lee, Sohn, Lee, Lewis, and
  Singh}]{carvalho2020reinforcement}
Carvalho, W.; Liang, A.; Lee, K.; Sohn, S.; Lee, H.; Lewis, R.~L.; and Singh,
  S. 2020.
\newblock Reinforcement Learning for Sparse-Reward Object-Interaction Tasks in
  First-person Simulated 3D Environments.
\newblock \emph{arXiv preprint arXiv:2010.15195}.

\bibitem[{Duan et~al.(2016)Duan, Schulman, Chen, Bartlett, Sutskever, and
  Abbeel}]{duan2016rl}
Duan, Y.; Schulman, J.; Chen, X.; Bartlett, P.~L.; Sutskever, I.; and Abbeel,
  P. 2016.
\newblock $\text{RL}^2$: Fast reinforcement learning via slow reinforcement
  learning.
\newblock \emph{arXiv preprint arXiv:1611.02779}.

\bibitem[{Erol(1996)}]{erol1996hierarchical}
Erol, K. 1996.
\newblock \emph{Hierarchical task network planning: formalization, analysis,
  and implementation}.
\newblock Ph.D. thesis.

\bibitem[{Fikes and Nilsson(1971)}]{fikes1971strips}
Fikes, R.~E.; and Nilsson, N.~J. 1971.
\newblock STRIPS: A new approach to the application of theorem proving to
  problem solving.
\newblock \emph{Artificial intelligence}, 2(3-4): 189--208.

\bibitem[{Finn, Abbeel, and Levine(2017)}]{finn2017model}
Finn, C.; Abbeel, P.; and Levine, S. 2017.
\newblock Model-agnostic meta-learning for fast adaptation of deep networks.
\newblock In \emph{International Conference on Machine Learning}, 1126--1135.
  PMLR.

\bibitem[{Fix(1985)}]{fix1985discriminatory}
Fix, E. 1985.
\newblock \emph{Discriminatory analysis: nonparametric discrimination,
  consistency properties}, volume~1.
\newblock USAF school of Aviation Medicine.

\bibitem[{Frank and J{\'o}nsson(2003)}]{frank2003constraint}
Frank, J.; and J{\'o}nsson, A. 2003.
\newblock Constraint-based attribute and interval planning.
\newblock \emph{Constraints}, 8(4): 339--364.

\bibitem[{Ghazanfari and Taylor(2017)}]{ghazanfari2017autonomous}
Ghazanfari, B.; and Taylor, M.~E. 2017.
\newblock Autonomous extracting a hierarchical structure of tasks in
  reinforcement learning and multi-task reinforcement learning.
\newblock \emph{arXiv preprint arXiv:1709.04579}.

\bibitem[{Huang et~al.(2019)Huang, Nair, Xu, Zhu, Garg, Fei-Fei, Savarese, and
  Niebles}]{huang2019neural}
Huang, D.-A.; Nair, S.; Xu, D.; Zhu, Y.; Garg, A.; Fei-Fei, L.; Savarese, S.;
  and Niebles, J.~C. 2019.
\newblock Neural task graphs: Generalizing to unseen tasks from a single video
  demonstration.
\newblock In \emph{Proceedings of the IEEE/CVF Conference on Computer Vision
  and Pattern Recognition}, 8565--8574.

\bibitem[{Kolve et~al.(2017)Kolve, Mottaghi, Han, VanderBilt, Weihs, Herrasti,
  Gordon, Zhu, Gupta, and Farhadi}]{ai2thor}
Kolve, E.; Mottaghi, R.; Han, W.; VanderBilt, E.; Weihs, L.; Herrasti, A.;
  Gordon, D.; Zhu, Y.; Gupta, A.; and Farhadi, A. 2017.
\newblock {AI2-THOR: An Interactive 3D Environment for Visual AI}.
\newblock \emph{arXiv}.

\bibitem[{Loula, Baroni, and Lake(2018)}]{loula2018rearranging}
Loula, J.; Baroni, M.; and Lake, B.~M. 2018.
\newblock Rearranging the familiar: Testing compositional generalization in
  recurrent networks.
\newblock \emph{arXiv preprint arXiv:1807.07545}.

\bibitem[{Luong, Pham, and Manning(2015)}]{luong2015effective}
Luong, M.-T.; Pham, H.; and Manning, C.~D. 2015.
\newblock Effective approaches to attention-based neural machine translation.
\newblock \emph{arXiv preprint arXiv:1508.04025}.

\bibitem[{Mehta, Tadepalli, and Fern(2011)}]{mehta2011autonomous}
Mehta, N.; Tadepalli, P.; and Fern, A. 2011.
\newblock Autonomous learning of action models for planning.
\newblock \emph{Advances in Neural Information Processing Systems}, 24:
  2465--2473.

\bibitem[{Muggleton and De~Raedt(1994)}]{muggleton1994inductive}
Muggleton, S.; and De~Raedt, L. 1994.
\newblock Inductive logic programming: Theory and methods.
\newblock \emph{The Journal of Logic Programming}, 19: 629--679.

\bibitem[{Oh et~al.(2017)Oh, Singh, Lee, and Kohli}]{oh2017zero}
Oh, J.; Singh, S.; Lee, H.; and Kohli, P. 2017.
\newblock Zero-shot task generalization with multi-task deep reinforcement
  learning.
\newblock In \emph{International Conference on Machine Learning}, 2661--2670.
  PMLR.

\bibitem[{Palatucci et~al.(2009)Palatucci, Pomerleau, Hinton, and
  Mitchell}]{Palatucci2009ZeroshotLW}
Palatucci, M.; Pomerleau, D.; Hinton, G.~E.; and Mitchell, T.~M. 2009.
\newblock Zero-shot Learning with Semantic Output Codes.
\newblock In \emph{NIPS}.

\bibitem[{Pennington, Socher, and Manning(2014)}]{pennington2014glove}
Pennington, J.; Socher, R.; and Manning, C.~D. 2014.
\newblock Glove: Global vectors for word representation.
\newblock In \emph{Proceedings of the 2014 conference on empirical methods in
  natural language processing (EMNLP)}, 1532--1543.

\bibitem[{Sohn, Oh, and Lee(2018)}]{sohn2018hierarchical}
Sohn, S.; Oh, J.; and Lee, H. 2018.
\newblock Hierarchical reinforcement learning for zero-shot generalization with
  subtask dependencies.
\newblock \emph{arXiv preprint arXiv:1807.07665}.

\bibitem[{Sohn et~al.(2020)Sohn, Woo, Choi, and Lee}]{sohn2020meta}
Sohn, S.; Woo, H.; Choi, J.; and Lee, H. 2020.
\newblock Meta reinforcement learning with autonomous inference of subtask
  dependencies.
\newblock \emph{arXiv preprint arXiv:2001.00248}.

\bibitem[{Su{\'a}rez-Hern{\'a}ndez et~al.(2020)Su{\'a}rez-Hern{\'a}ndez,
  Segovia-Aguas, Torras, and Aleny{\`a}}]{suarez2020strips}
Su{\'a}rez-Hern{\'a}ndez, A.; Segovia-Aguas, J.; Torras, C.; and Aleny{\`a}, G.
  2020.
\newblock Strips action discovery.
\newblock \emph{arXiv preprint arXiv:2001.11457}.

\bibitem[{Sutton, Precup, and Singh(1999)}]{sutton1999between}
Sutton, R.~S.; Precup, D.; and Singh, S. 1999.
\newblock Between MDPs and semi-MDPs: A framework for temporal abstraction in
  reinforcement learning.
\newblock \emph{Artificial intelligence}, 112(1-2): 181--211.

\bibitem[{Walsh and Littman(2008)}]{walsh2008efficient}
Walsh, T.~J.; and Littman, M.~L. 2008.
\newblock Efficient learning of action schemas and web-service descriptions.
\newblock In \emph{AAAI}, volume~8, 714--719.

\bibitem[{Xu et~al.(2018)Xu, Nair, Zhu, Gao, Garg, Fei-Fei, and
  Savarese}]{xu2018neural}
Xu, D.; Nair, S.; Zhu, Y.; Gao, J.; Garg, A.; Fei-Fei, L.; and Savarese, S.
  2018.
\newblock Neural task programming: Learning to generalize across hierarchical
  tasks.
\newblock In \emph{2018 IEEE International Conference on Robotics and
  Automation (ICRA)}, 3795--3802. IEEE.

\bibitem[{Zhuo et~al.(2010)Zhuo, Yang, Hu, and Li}]{zhuo2010learning}
Zhuo, H.~H.; Yang, Q.; Hu, D.~H.; and Li, L. 2010.
\newblock Learning complex action models with quantifiers and logical
  implications.
\newblock \emph{Artificial Intelligence}, 174(18): 1540--1569.

\end{thebibliography}

\newpage
\appendix
\section{Appendix}


\begin{figure*}[h!]
  \centering
  \includegraphics[trim=10 50 10 30,clip,width=0.8\textwidth]{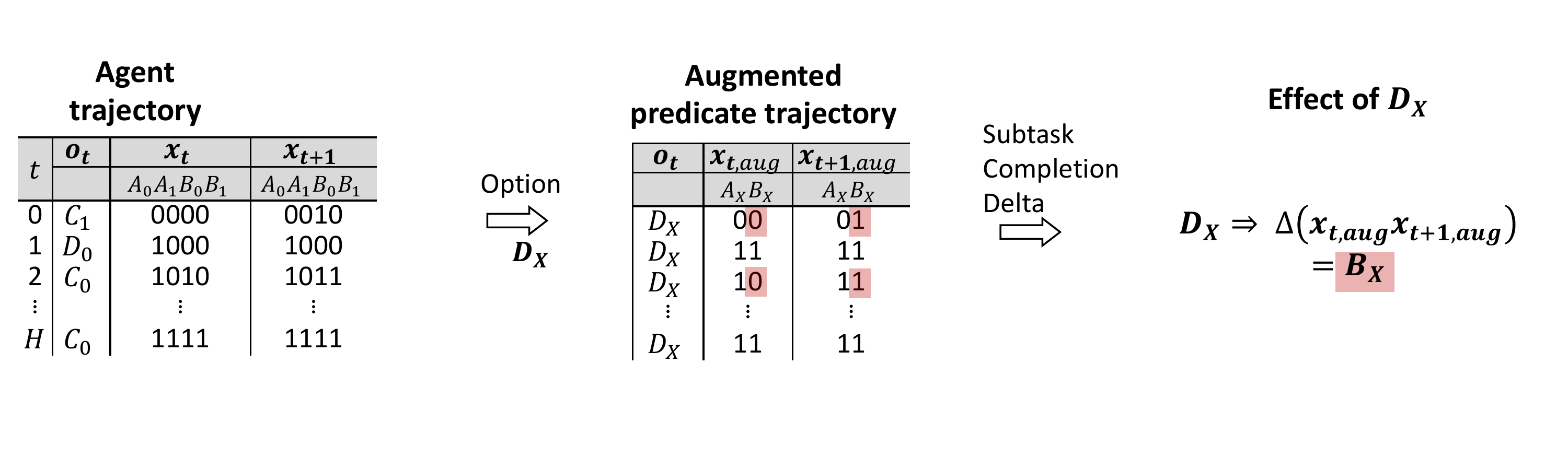}
  \caption{An overview of our approach for estimating the \textbf{parameterized effect} of the parameterized subtask graph, $\widehat{\FGeffect}$, in a simple environment with subtasks $A, B$ and options $C, D, E$.
  Each subtask and option has a parameter 0 or 1.
  We run effect inference for every option and show $D_X$ as an example.
  \textbf{1.} The first table is built from the agent's trajectory ($x_t, o_t$ is the subtask
  completion and option executed at time $t$).
  \textbf{2.} We build the second table, the ``augmented'' trajectory by substituting
  $X$ into all possible subtask completions, $A_X, B_X$, and restricting the table to
  only row where $o_t = D_X$.
  \textbf{3.} We infer option dynamics, $(x_t, o_t) \mapsto x_{t+1}$, by calculating the simple aggregated difference between subtask completion
  before and after $D_X$, $\Delta(x_{t,\text{aug}}, x_{t+1,\text{aug}})$.
  }
  \label{fig:pilp-effect-overview}
\end{figure*}

\section{Details on the Method}
\label{app:methods}

\subsection{Parameterized Subtask Graph MLE Derivation}
Recall in the Parameterized Subtask Graph Inference section our goal is to
infer the maximum likelihood factored subtask graph
$\FG$ given a trajectory $\tau_H$.
\begin{equation}
    \widehat{\FG}^\text{MLE} = \argmax_{\FGpcond, \FGeffect, \FGr} p(\tau_H | \FGpcond, \FGeffect, \FGr)
\end{equation}
where
$\tau_H = \{ s_1, o_1, r_1, d_1, \dots, s_H\}$ is the adaptation trajectory
of the adaptation policy $\pi_\theta^{\text{adapt}}$ after $H$ time steps.

We can expand the likelihood term as:
\begin{align}
    &p(\tau_H | \FGpcond, \FGeffect, \FGr)
    \\=&\; p(s | \FGpcond, \FGeffect)
    \prod_{t=1}^H
    \Big[
    \pi_\theta (o_t | \tau_t)
    \\&\;\;
    p(s_{t+1} | s_t, o_t, \FGpcond, \FGeffect)
    p(r_t | s_t, o_t, \FGr)
    p(d_t | s_t, o_t)
    \Big]
    \\\propto&\; p(s | \FGpcond, \FGeffect)
    \prod_{t=1}^H
    \Big[
    p(s_{t+1} | s_t, o_t, \FGpcond, \FGeffect)
    \\&\;\;
    p(r_t | s_t, o_t, \FGeffect, \FGr)
    \Big]
\end{align}
where we dropped terms independent of $\FG$.
From our definitions of the Parameterized Subtask Graph problem,
the predicate precondition $\FGpcond$ determines the mapping from completion $x$ to
option eligibility $e$, $x \mapsto e$,
the predicate effect $\FGeffect$ determines the mapping from completion and option
to completion, $(x_t, o) \mapsto x_{t+1}$,
and finally, the predicate reward $\FGr$ determines the reward given
when a subtask is completed at time $t$.
Then, we can rewrite the MLE as:
\begin{align}
    \widehat{PG}^\text{MLE}
    =& \left(\hatFGpcond^\text{MLE}, \hatFGeffect^\text{MLE}, \hatFGr^\text{MLE}\right)
    \\=& \bigg(
    \argmax_{\FGpcond}\prod_{t=1}^H p(e_t | x_t, \FGpcond),
    \\&
    \argmax_{\FGeffect}\prod_{t=1}^H p(x_{t+1} | x_t, o_t, \FGeffect),\label{eq:mle-exp-2}
    \\&\argmax_{\FGr}\prod_{t=1}^H p(r_t | o_t, o_{t+1}, \FGr)\label{eq:mle-exp-3}
    \bigg)
\end{align}

The rest of PSGI follows in maximizing $\hatFGpcond$, $\hatFGeffect$, and $\hatFGr$ individually.

\begin{figure*}[!h]
  \centering
  \includegraphics[trim=15 25 10 38,clip,width=1.0\textwidth]
  {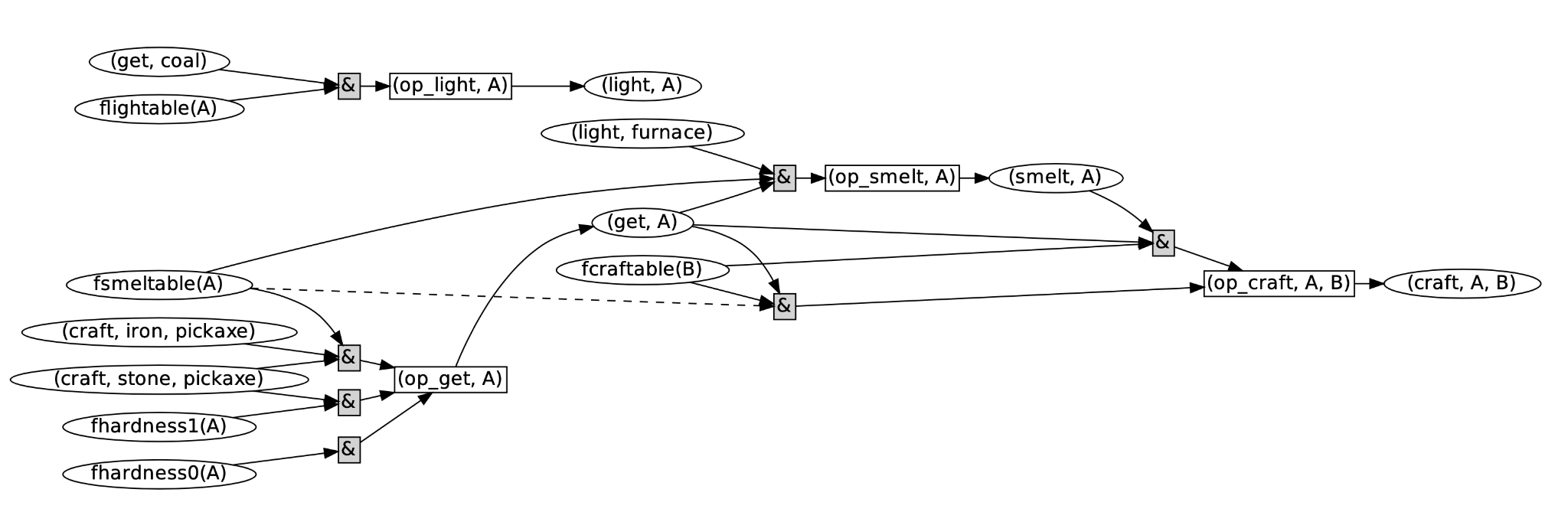}
  \caption{The inferred parameterized subtask graph by \fsgibf after 6000 timesteps in the \mine domain.
  Options are represented in rectangular nodes. Subtask completions and attributes
  are are in oval nodes. A solid line represents a positive precondition / effect, dashed for negative.
  Ground truth attributes are included option/subtask parameters, however
  which attributes are used for which option preconditions is still hidden, which
  PSGI must infer.
  }
  \label{fig:etmining-pilp}
\end{figure*}
\section{Details on the Tasks}
\label{app:task}

\paragraph{AI2Thor} The \thor domain is modelled after interactions from
the AI2Thor simulated home environment~\cite{ai2thor}.
The agent has 4 main interactions in the environment: \texttt{pickup},
\texttt{put}, \texttt{slice}, and \texttt{cook}. The agent can execute options to move objects
with \texttt{pickup} and \texttt{put}, and execute options to change object states
with \texttt{slice} and \texttt{cook}.
Figure~\ref{fig:psgi-overview} shows a simplified version of the preconditions and
effects of interactions.
All attributes of objects from~\citep{ai2thor} were maintained in our simulated \thor domain,
such as the pickup-receptacle compatibility properties (e.g. a potato cannot be placed on the toilet),
and object transformation properties (e.g. a potato can be cooked but lettuce cannot).
There are 39 unique object types present in \thor, each with respective pickup-receptacle compatibilities
and transformative properties.
To make the \thor domain more difficult and test more on generalization, we added 13
additional food object types. These new food object types were coded to have similar attributes
to the existing foods, but have different appearance/embeddings. E.g. A new object \textit{Yam}
was added, with similar properties to the potato.
\paragraph{Cooking} The \cook domain is also modelled after cooking interactions from
the AI2Thor simulated home environment~\cite{ai2thor}.
However, \cook is much simpler than \thor. Notably, the pickup-receptacle properties
are simplified to objects either being pickupable, or placeable.
Similarly, the agent has 4 main interactions in the environment: \texttt{pickup},
\texttt{put}, \texttt{slice}, and \texttt{cook}.
Figure~\ref{fig:psgi-overview} shows a simplified version of the preconditions and
effects of interactions. The full parameterized subtask graph is shown in
Figure~\ref{fig:cooking-pilp}, which was correctly inferred by the PSGI agent.
\paragraph{Mining} The \mine domain is modelled after the open world video game Minecraft
and the domain introduced by~\cite{sohn2018hierarchical}.
Similar to the mining from~\cite{sohn2018hierarchical}, the agent has
4 main interactions in the environment:
\texttt{get}, \texttt{light}, \texttt{smelt}, and \texttt{craft}.
The agent may retrieve/mine materials with \texttt{get}, use \texttt{light}
and \texttt{smelt} to prepare materials in order to \texttt{craft} them into
usable tools.
However, \mine has additional added complexity from~\cite{sohn2018hierarchical}.
\mine has one more ``tier'' of mining difficulty --- a stone pickaxe must be used
to mine iron, and a iron pickaxe must be used to mine diamond.
This makes \mine significantly more difficult for the agent, and more closely
matches the gameplay in the Minecraft video game.
\mine also has many more materials added than~\cite{sohn2018hierarchical}.
E.g.\ in some tasks the agent will encounter \texttt{stone}, \texttt{iron}, \texttt{copper},
\texttt{gold}, \texttt{diamond}, etc, each material with similar or different latent attributes.
Again, this makes the task more difficult, but more similar to Minecraft.
The latent parameterized subtask graph of \mine is shown in Figure~\ref{fig:etmining-pilp}.

\section{Details on the Baselines and Hyperparameters}
\label{app:baseline}

\paragraph{HRL} The baseline \hrl is an actor-critic model over pre-learned options
~\cite{andreas2017modular}.
As our compared approach \fsgibf utilizes the \textit{entity embeddings} (used for zero-shot
learning entity attributes), we use an architecture for \hrl that uses attention
over the entities as well as the observations, following \cite{luong2015effective}.
We briefly describe this architecture.

Let $x \in \{0, 1\}^N, e \in \{0, 1\}^M$ be the completion and eligibility vector respectively,
where there are $N$ subtasks and $M$ options.
Let $\vect{E_x} \in \mathbb{R}^{N, D}, \vect{E_e} \in \mathbb{R}^{N, D}$ be the entity embeddings for each
subtask and option entities concatenated.

Let $D'$ be the embedding dimension. We apply attention over the observations by:
\begin{align*}
    \vect{V} &= [\vect{E_x} ; \vect{E_e}] \vect{W_V} [x ; e] \in \mathbb{R}^{N+M, D'}
    \\ \vect{K} &= [\vect{E_x} ; \vect{E_e}] \vect{W_K} [x ; e] \in \mathbb{R}^{N+M, D'}
    \\ \text{attention} &= \vect{V} \text{softmax}(\vect{W_Q} \vect{K})
\end{align*}

Similarly, we also use attention to calculate the option logits:
\begin{align*}
    \vect{h} &= \text{MLP}(\text{attention} ; \text{observation}) \in \mathbb{R}^{D'}
    \\ \vect{O} &= \vect{W_O} \vect{E_e} \in \mathbb{R}^{M, D'}
    \\ \text{logits} &= \vect{O} \vect{h}
\end{align*}

We searched through the following hyperparameters for \hrl in a simple grid search.

\begin{center}
\begin{tabular}{l|r}
    \hrl hyperparameters\\
    \hline
    Learning Rate & \{1e-4, 2.5e-4\} \\
    Entropy Cost & \{0.01, 0.03\} \\
    Baseline Cost & 0.5 \\
    $N$-step horizon & 4 \\
    Discount & 0.99
\end{tabular}
\end{center}

\paragraph{MSGI+}
We implement \msgi following work from~\cite{sohn2020meta}, however, we adjust
prior work to additionally infer option effects (where previous options were assumed
to only complete singular subtasks), by using the \fsgibf effect model, but without
leveraging the smoothness assumptions. I.e.\ we directly infer effect from ~\Cref{eq:non-smooth-effect}, or, skipping step \textbf{2.} of ~\Cref{fig:pilp-effect-overview}.

We use the following hyperparameters for \msgi.
\begin{center}
\begin{tabular}{l|r}
    \msgi hyperparameters\\
    \hline
    Exploration & count-based \\
    GRProp Temperature & 200.0 \\
    GRProp $w_a$ & 3.0 \\
    GRProp $\beta_a$ & 8.0 \\
    GRProp $\epsilon_\text{or}$ & 0.8 \\
    GRProp $t_\text{or}$ & 2.0
\end{tabular}
\end{center}

\paragraph{PSGI}
We use the following hyperparameters for \fsgibf. We use mostly similar parameters to \msgi.
\begin{center}
\begin{tabular}{l|r}
    \fsgibf hyperparameters\\
    \hline
    Exploration & count-based \\
    GRProp Temperature & 200.0 \\
    GRProp $w_a$ & 3.0 \\
    GRProp $\beta_a$ & 8.0 \\
    GRProp $\epsilon_\text{or}$ & 0.8 \\
    GRProp $t_\text{or}$ & 2.0 \\
    Number of priors & 4 \\
    Prior timesteps $T_\text{prior}$ & 2000
\end{tabular}
\end{center}

\section{Additional Experiments}
\label{app:experiments}

We additionally wanted to ask the following research questions:
\begin{enumerate}
    \item Can PSGI \textit{generalize} to unseen subtasks with unseen entities using
the inferred attributes?
    \item How does the quality of the PSGI's prior affect transfer learning performance?
\end{enumerate}

\begin{figure*}[t!]
  \centering
  \includegraphics[width=0.30\textwidth]{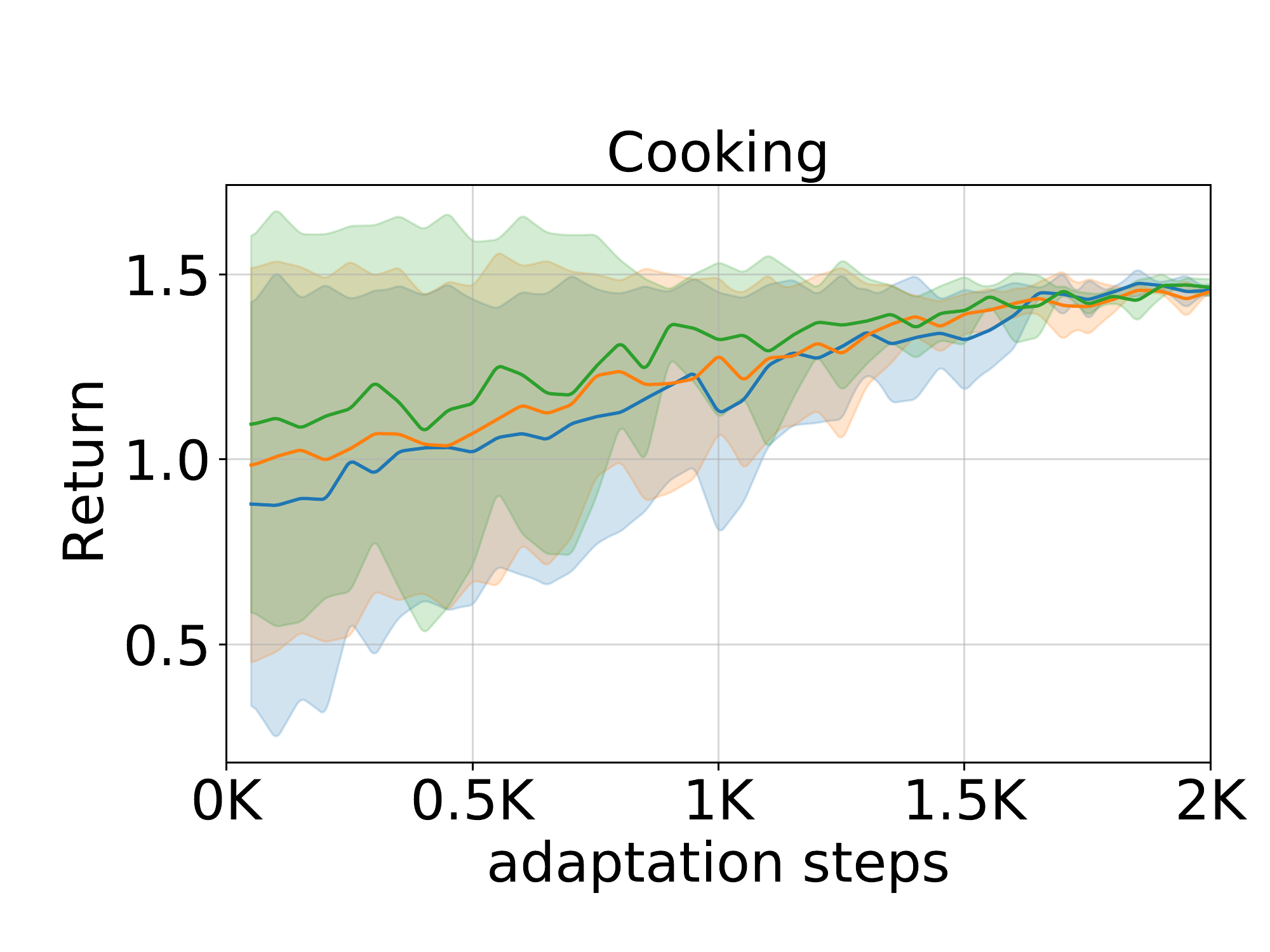}
  \includegraphics[width=0.30\textwidth]{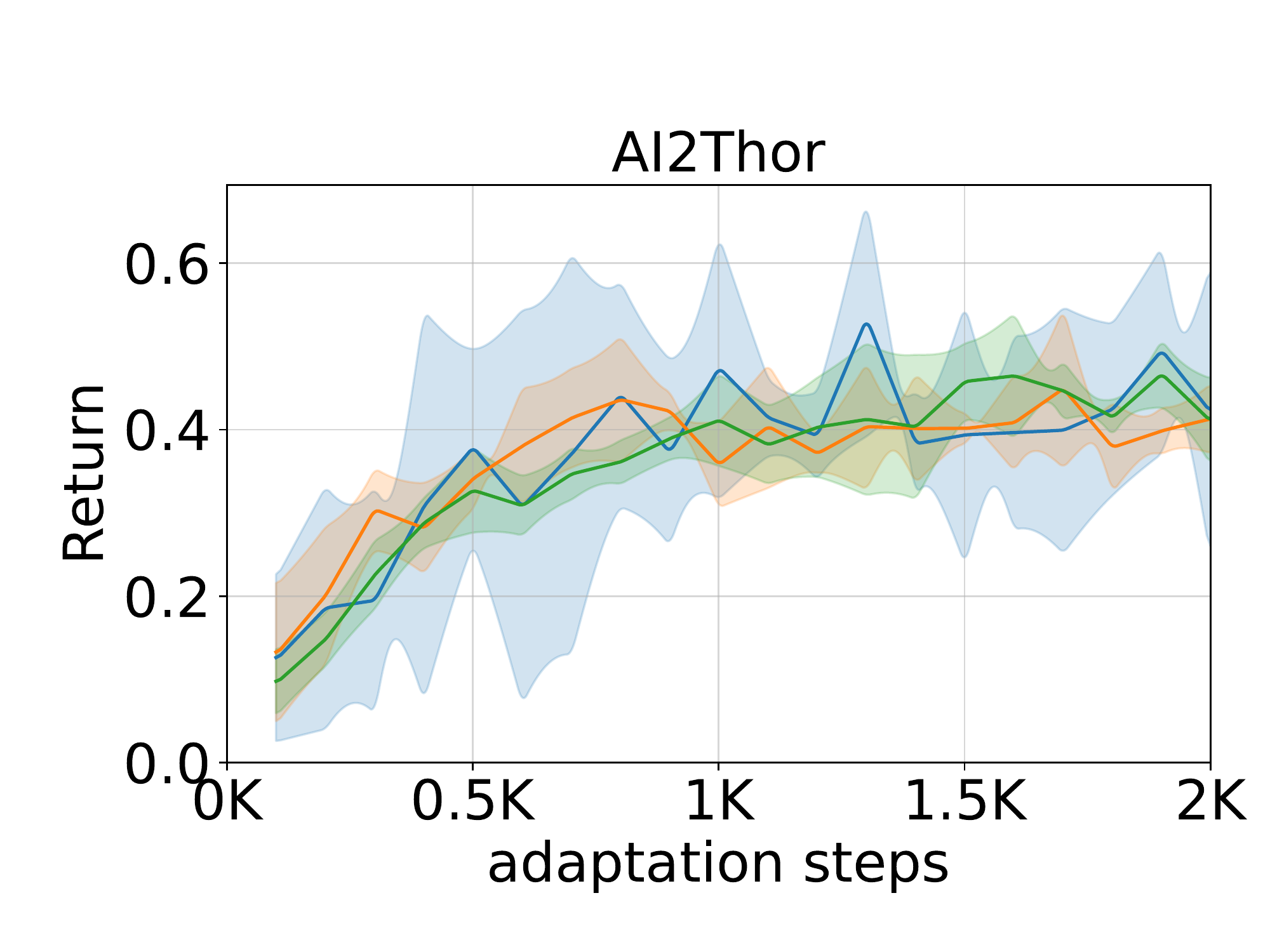}
  \includegraphics[width=0.30\textwidth]{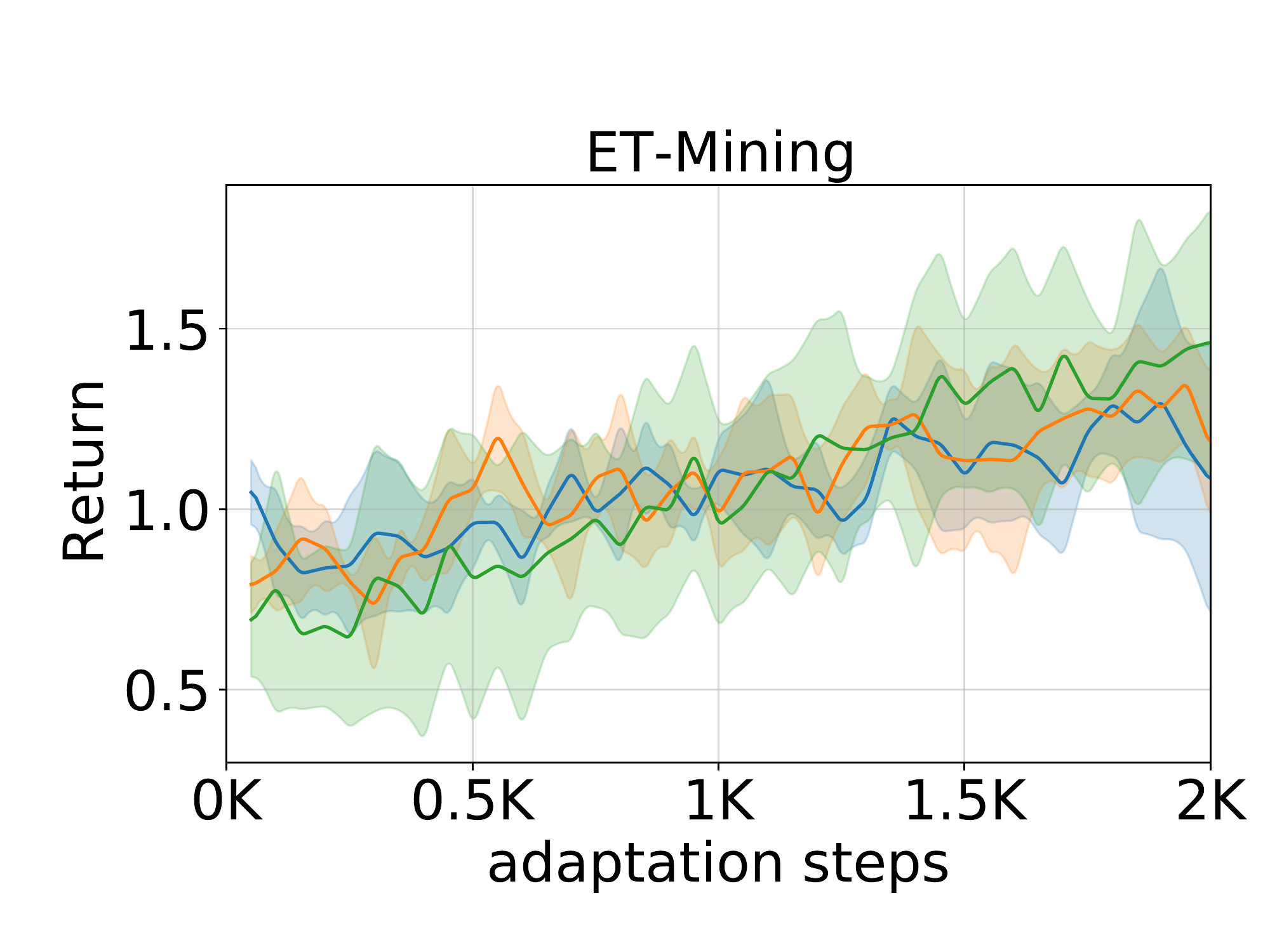}
  \\
  \includegraphics[width=0.30\textwidth]{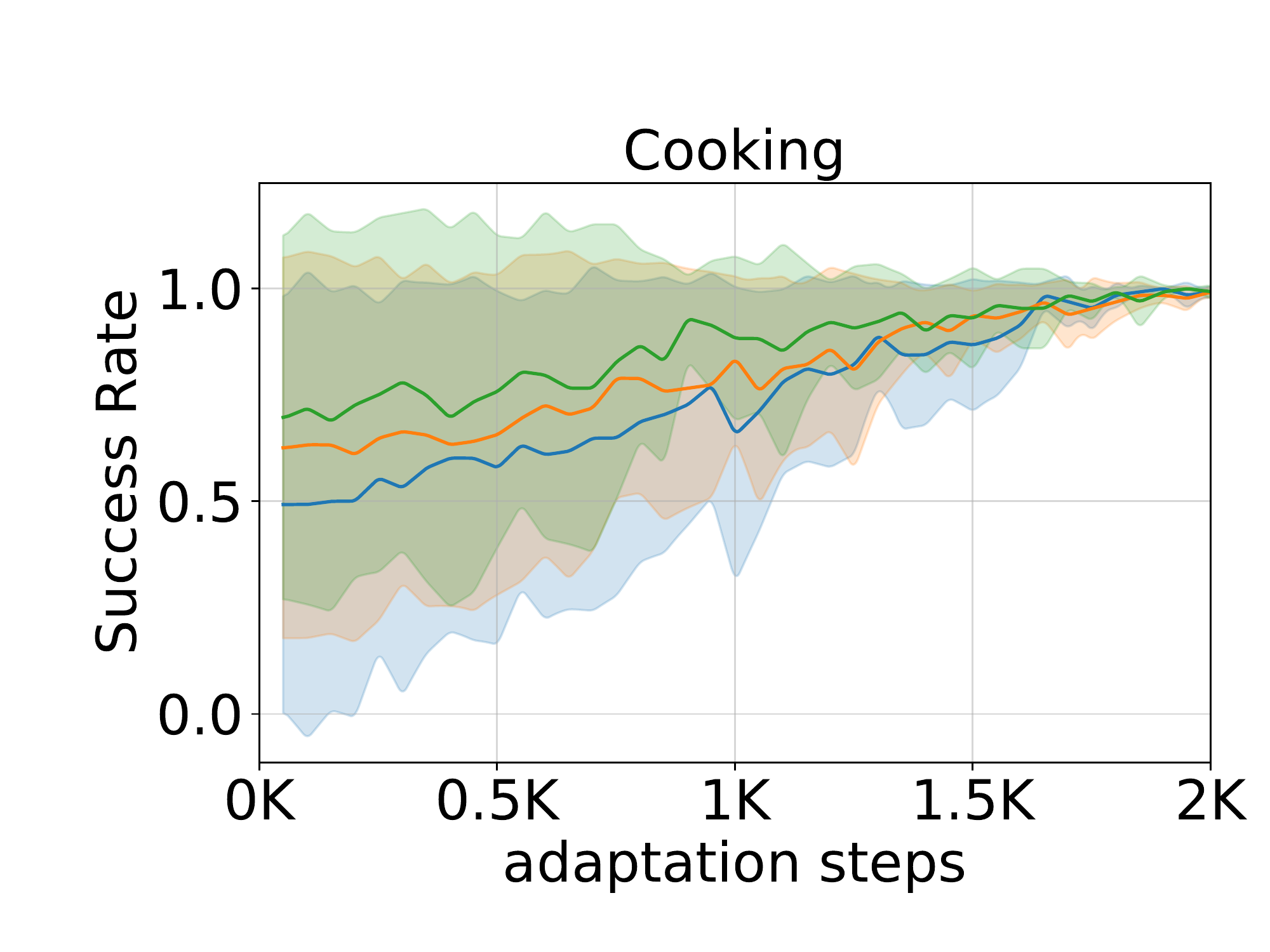}
  \includegraphics[width=0.30\textwidth]{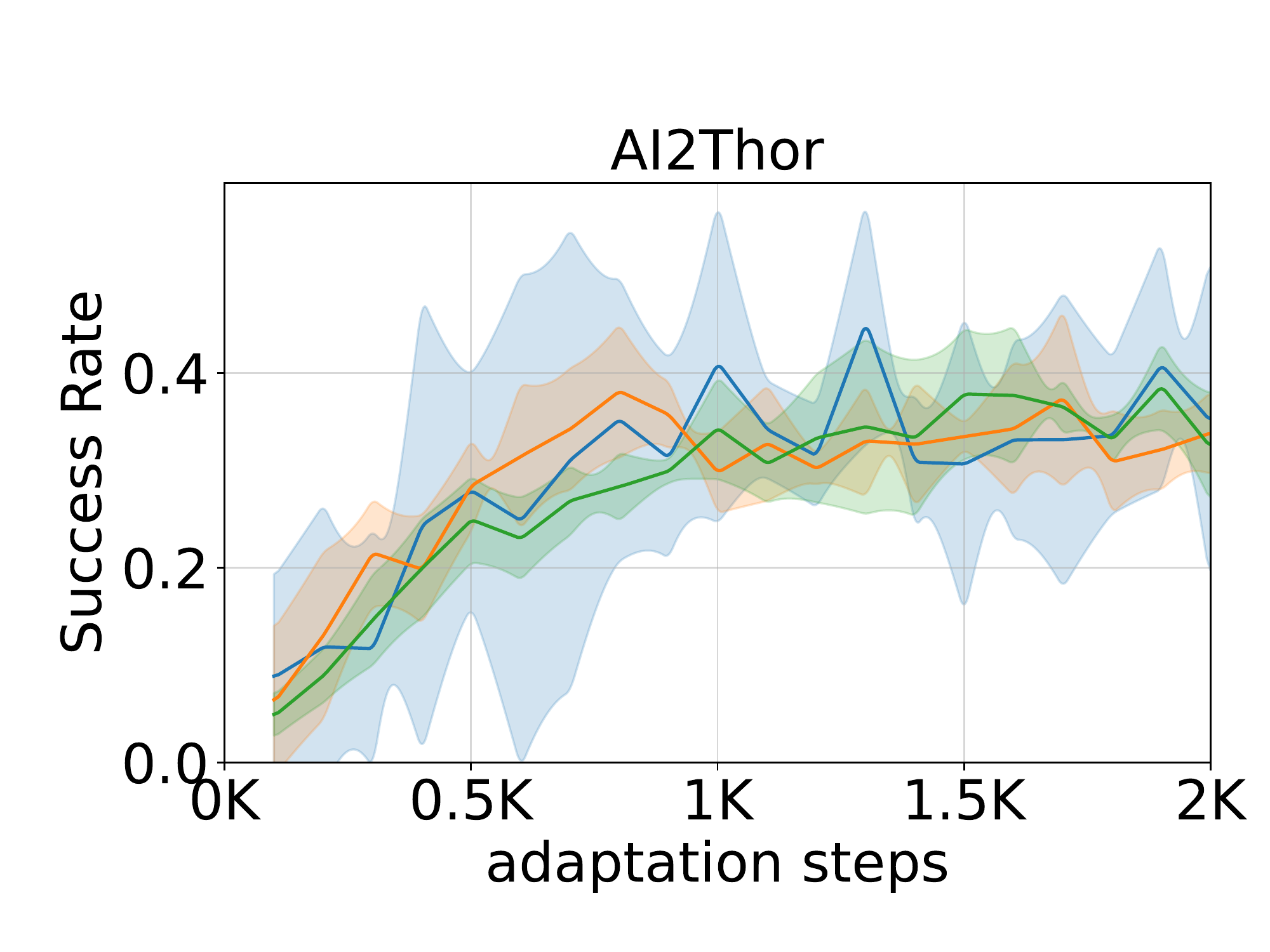}
  \includegraphics[width=0.30\textwidth]{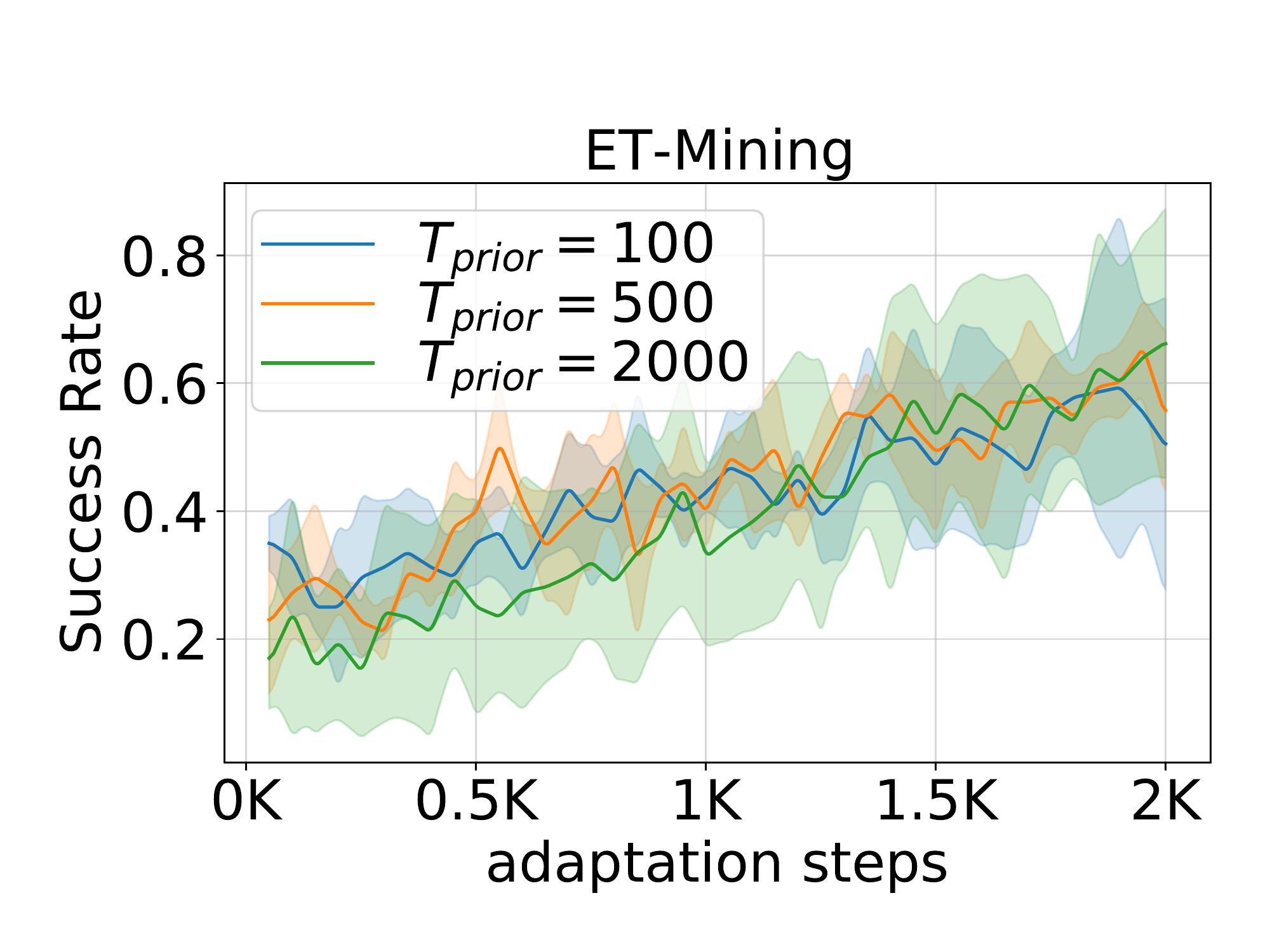}
  \caption{The adaptation curve of \fsgibf trained with different priors. Priors were
  trained in the training tasks of \cook, \thor, and \mine domains respectively
  for $T_\text{prior}= 100, 500,$ and $2000$ timesteps.
  }
  \label{fig:ablation_prior}
\end{figure*}

\subsection{Zero Shot Learning Attributes}
In our experiments, we assume the first entity of every subtask and option
serves as a ``verb'' entity (e.g. \texttt{pickup}, \texttt{cook}, etc.).
We assume there is there is no shared structure across subtasks and options with
different verbs.

As described in section~\ref{sec:zero-shot-learning-attributes},
(when attributes are not provided), we infer attributes from an exhaustive
powerset of all possible features on seen parameters.
The attributes that are used for the graph are then likely to be semantically
meaningful, as the decision tree selects the most efficient features.
Hence, to test whether PSGI is generalizable, we can evaluate \textit{whether
attributes are accurately inferred} for unseen parameters when only given the ground truth attributes
on seen parameters (given that PSGI will infer the ground truth for the seen parameters).

We measure the generalization error of PSGI if some ``weak" signal is provided
through parameters. We suppose the word labels for options and subtasks are provided in \cook.
I.e.\ the words for parameters ``pickup", ``apple", etc. are known.
Then, we can infer low level (but semantically meaningful) features from these words
by using \textit{word embeddings} to encode the parameters~\citep{pennington2014glove}.
We choose to use 50 dimensional GloVE word embeddings from~\citet{pennington2014glove}.
We then evaluate by measuring the accuracy of attributes for \textbf{20} additional
unseen test parameters, all words related to kitchens and cooking. We show the
results in Table~\ref{tab:attribute-inf-accuracy}.


\begin{table}[]
    \centering
    \begin{tabular}{l|r}
        Environment &
        \begin{tabular}{@{}c@{}}Attribute Generalization \\ Accuracy\end{tabular}\\
        \hline
        \cook & $94.9 \pm 1.0$\%\\
        \mine & $87.9 \pm 1.5$\% \\
        \thor & $82.3 \pm 3.9$\%
    \end{tabular}
    \caption{We evaluate the generalization accuracy of PSGI on unseen test entities
    in each environment. For each ground truth attribute, we evaluate
    whether PSGI accurately labels the unseen test entity correctly.
    We show the average accuracy over each ground truth attribute in environments.
    }
    \label{tab:attribute-inf-accuracy}
\end{table}
From these results, we can extrapolate that at least \textbf{70\%} of edges
(on unseen entities) in the predicate subtask graph using these attributes are accurate.

\subsection{Effect of the Prior}
As described in the Task Transfer and Adaptation section,
in PSGI, the training phase is used to learn a prior parameterized subtask
graph, $\widehat{\FG}_\text{prior}$.
$\widehat{\FG}_\text{prior}$ is then used in an ensemble with the the test policy's
learned graph $\widehat{\FG}_\text{test}$.
Then, we can infer that the more accurate $\widehat{\FG}_\text{prior}$ is, the better
the transfer policy will perform.
To study this, we varied the number of timesteps used to learn the priors
in the \cook, \thor, and \mine domains. We trained priors $T_\text{prior}= 100, 500,$ and $2000$ timesteps
respectively, shown in ~\Cref{fig:ablation_prior}.

In the \cook domain, we can see that when $T_\text{prior}$ is higher,
the average return and success rate increases.
However, in the \mine and \thor domain, we see no obvious correlation between $T_\text{prior}$ and performance.
We reason that this may be from a number of factors --- the prior graphs may not have
significant difference between $T_\text{prior} = 100, 500,$ and $2000$ timesteps,
or that $\widehat{\FG}_\text{prior}$ is significantly different from the latent
parameterized subtask graph during testing, rendering the prior less useful.

\begin{figure*}[h!]
  \centering
  \includegraphics[width=0.85\textwidth]{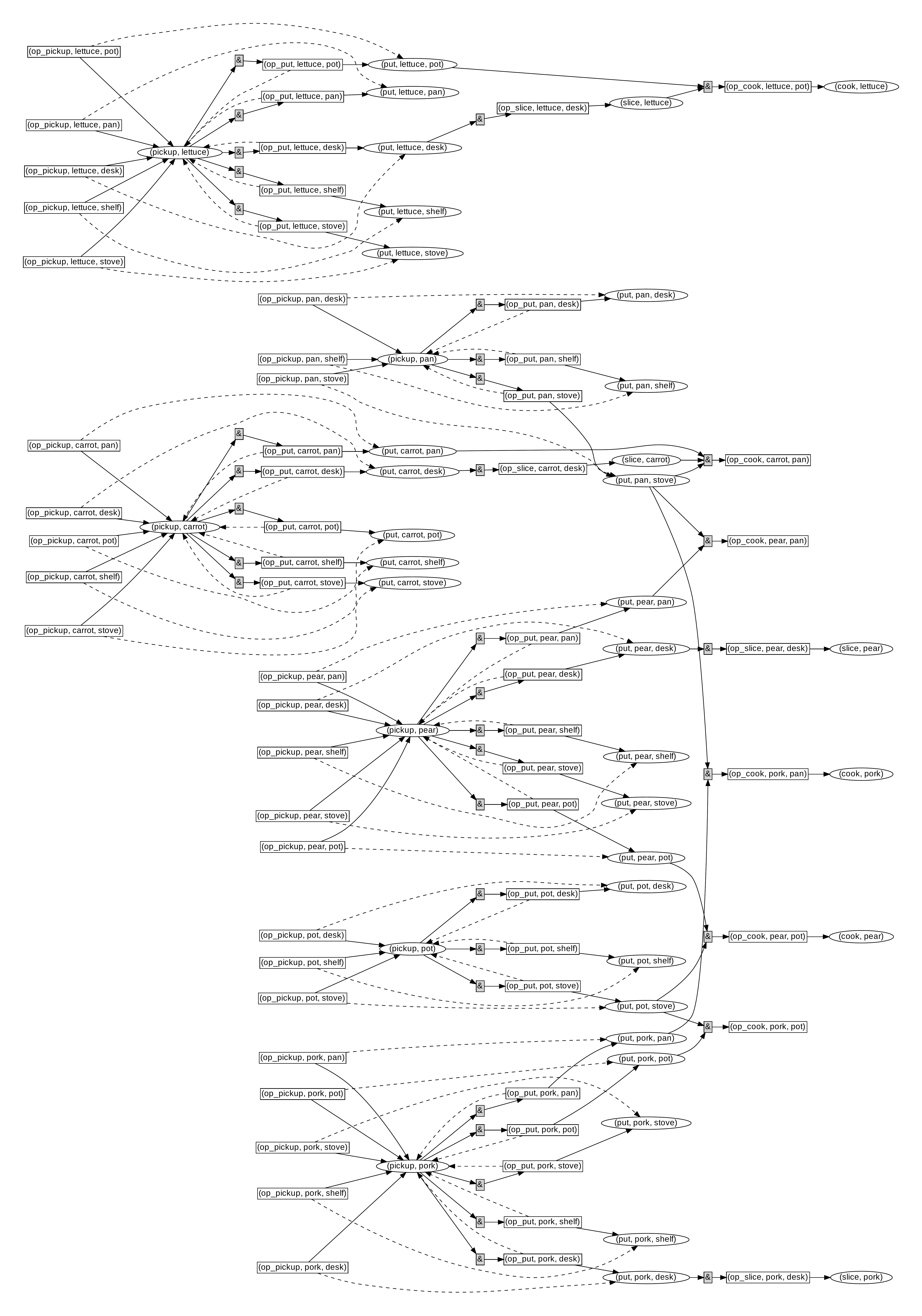}\vspace{-20pt}
  \caption{Inferred subtask graph  by \msgi after 2000 timesteps in the \cook environment.
  For \msgi, 262 options with no inferred precondition and effect were not visualized for readability.
  Options are represented in rectangular nodes. Subtask completions and attributes
  are are in oval nodes. A solid line represents a positive precondition / effect, dashed for negative.
  Ground truth attributes are included option/subtask parameters, however
  which attributes are used for which option preconditions is still hidden, which
  \msgi must infer.
  }
  \label{fig:cooking-pilp-msgi}
\end{figure*}

\end{document}